\documentclass[final,1p,times]{elsarticle}
\usepackage{booktabs, makecell, tabularx}
\usepackage{hyperref}       
\usepackage{url}            
\usepackage{graphicx}
\usepackage{amsmath}
\usepackage{pgfplots}
\usepackage[center]{caption}
\usepackage{subcaption}
\usepackage{multirow}   
\usepackage{longtable}
\usepackage{rotating}
\usepackage{pgfplots}
\pgfplotsset{width=10cm,compat=1.9}

\makeatletter
\def\ps@pprintTitle{%
    \let\@oddhead\@empty
    \let\@evenhead\@empty
    \let\@oddfoot\@empty
    \let\@evenfoot\@empty
}
\makeatother

\usepackage{amssymb}

\begin{document}

\begin{frontmatter}



\title{Random Heterogeneous Neurochaos Learning Architecture for Data Classification}

\author[inst1]{Remya Ajai A S\footnote{remya@am.amrita.edu}}

\affiliation[inst1]{organization={Department of Electronics and Communication Engineering},
            addressline={Amrita Vishwa Vidyapeetham}, 
            city={Amritapuri},
            country={India}}
  
\author[inst2]{Nithin Nagaraj\footnote{nithin@nias.res.in}}

\affiliation[inst2]{organization={Complex Systems Programme},
            addressline={National Institute of Advanced Studies}, 
            city={Indian Institute of Science Campus,~Bengaluru},
            state={Karnataka},
            country={India}}
\begin{abstract}
 Inspired by the human brain's structure and function, Artificial Neural Networks (ANN) were developed for data classification. However, existing Neural Networks, including Deep Neural Networks, do not mimic the brain's rich structure. They lack key features such as randomness and neuron heterogeneity, which are inherently chaotic in their firing behavior. Neurochaos Learning (NL), a chaos-based neural network, recently employed one-dimensional chaotic maps like Generalized Lüroth Series (GLS) and Logistic map as neurons. For the first time, we propose a random heterogeneous extension of NL, where various chaotic neurons are randomly placed in the input layer, mimicking the randomness and heterogeneous nature of human brain networks. We evaluated the performance of the newly proposed Random Heterogeneous Neurochaos Learning (RHNL) architectures combined with traditional Machine Learning (ML) methods. On public datasets, RHNL outperformed both homogeneous NL and fixed heterogeneous NL architectures in nearly all classification tasks. RHNL achieved high F1 scores on the Wine dataset (1.0), Bank Note Authentication dataset (0.99), Breast Cancer Wisconsin dataset (0.99), and Free Spoken Digit Dataset (FSDD) (0.98). These RHNL results are among the best in the literature for these datasets. We investigated RHNL performance on image datasets, where it outperformed stand-alone ML classifiers. In low training sample regimes, RHNL was the best among stand-alone ML. Our architecture bridges the gap between existing ANN architectures and the human brain's chaotic, random, and heterogeneous properties. We foresee the development of several novel learning algorithms centered around Random Heterogeneous Neurochaos Learning in the coming days.

\end{abstract}









\end{frontmatter}

\section{Introduction}
\label{sec:sample1}
Brain consists of complex networks of enormous number of neurons which are inherently non-linear\cite{ramachandran1998phantoms}. Inspired by the human brain in the way biological neurons are signaling to one another, Artificial Neural Networks (ANN) were developed for purposes of information processing and classification. With the rapid growth of Artificial Intelligence (AI) algorithms and easy availability of highly efficient and inexpensive computing hardware, almost all application domains today utilize Machine Learning (ML) algorithms/techniques and Deep Learning (DL) architectures for various tasks. Applications of AI include (and not limited to) speech processing~\cite{graves2013speech},  cybersecurity~\cite{harikrishnan2018machine}, computer vision~\cite{sebe2005machine}, and medical diagnosis~\cite{harikrishnan2019vision,remya2020analysis,krishna2019analysis,asif2023enhanced}. There are also algorithms that were developed to relate with the human brain in terms of learning and memory encoding~\cite{aihara1990chaotic}.  The learning algorithms perform internal weight updates and optimize their hyperparameter values so that error functions are minimised. Even though ANN is a huge success today, it needs to be greatly improved in order to mimic the human brain in terms of energy-efficient performance of complex tasks. Thus, in recent times, there has been a focus towards developing novel biologically-inspired algorithms and learning architectures by various researchers~\cite{delahunt2019putting, balakrishnan2019chaosnet}. 

\begin{table}[h!]
\centering
\caption{A brief comparison of ANNs and biological neural networks. 
\label{table0:ann_bnn}}
\begin{tabular}{|p{6cm}|p{6cm}|}
\hline
\textbf{Artificial Neural Networks (ANN)}             & \textbf{Biological Neural Networks} \\ 
\hline
\includegraphics[scale=0.5]{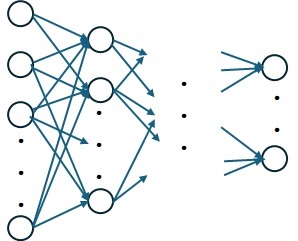} 
  		&
\includegraphics[scale=0.5]{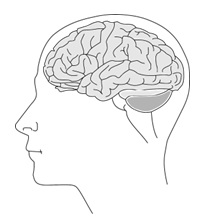}\\
\hline
Homogeneous neurons & Heterogeneous neurons \\
\hline
Every neuron performs a weighted linear combination of inputs followed by nonlinear activation & Neurons classified based on structure (unipolar, bipolar, pseudounipolar or multipolar) and function (sensory, motor, interneurons) \\
\hline
Scalar valued output at every neuron & Vector valued output as different neurons fire at different rates and duration \\
\hline
Non-chaotic neurons & Chaotic neurons (spiking and bursting behaviour) \\
\hline
Complexity of network of neurons is achieved through depth & Complexity achieved through depth, heterogeneity, randomness and differentiated processing \\
\hline
Causal structures of input dataset is not preserved internally in ANNs~\cite{nb2022causality}  & Internal representation of input stimuli preserves causal structures~\cite{nb2022causality} \\
\hline
\end{tabular}
\end{table}

There are fundamental differences between the way the human brain functions at the level of a single neuron (the basic unit that receives, processes and transmits information) and how ANN processes information (see Table~\ref{table0:ann_bnn}). Brain neurons are found to exhibit chaotic behaviour~\cite{korn2003there} whereas neurons in ANNs perform simple weighted addition of input data. Typically, in existing ANNs, homogeneous neurons are used whereas in the central nervous system/brain~\cite{perez2021neural}, neurons are known to be heterogeneous. These neurons in biological neural networks are differentiated based on structure and function. Sensory neurons are activated by input of sensory stimuli from the external environment. Motor neurons of the spinal cord connect to the muscle glands and organs throughout the human body. Interneurons connect spinal motor and sensory neurons. Based on structure of the neurons, they are classified as either being unipolar, bipolar, pseudounipolar or multipolar~\cite{Weis2019}. Based on these observations, we can correctly say that current ANNs are only loosely inspired by the brain.

Neurochaos Learning (or NL) is a recently developed brain-inspired chaos-based artificial neural network for data classification~\cite{balakrishnan2019chaosnet, harikrishnan2020neurochaos}. Majority of Machine Learning (ML) algorithms have relied heavily on substantial datasets for acquiring knowledge about the underlying distribution.  The first of NL architectures, dubbed  \verb|ChaosNet|, has demonstrated state-of-the-art performance in classification tasks with only a fraction of training samples needed for learning. Subsequently, NL was shown to perform equally well on imbalanced datasets, as well as, boost the performance of standard ML classifiers (SVM, kNN and others)~\cite{sethi2023neurochaos}. Moderate levels of noise within the context of neurochaos learning are found to optimize performance in classification tasks~\cite{harikrishnan2021noise}. It is no surprise that NL has found to preserve causal structures of input dataset in its internal representation of chaotic neural traces~\cite{nb2022causality} which is completely missing in the internal representation of ANNs.   

In our previous study~\cite{as2023analysis}, we have proposed an extension of NL architecture to incorporate heterogeneous neurons. We first demonstrated that the NL architecture with homogeneous neurons, but with a different 1D chaotic map (the logistic map) than the one used in \verb|ChaosNet| (1D Generalized L\"{u}roth Series or GLS map) is also equally good at learning tasks. Classification accuracies for {\it Ionosphere, Statlog (Heart), Bank Note Authentication, Breast Cancer Wisconsin, Haberman's Survival} and {\it Seeds} increased with the use of one-dimensional (1D) logistic map (with chaotic behaviour regime) as neurons compared with GLS maps as neurons. We then proposed HNL: Heterogeneous Neurochaos Learning which combined GLS maps as neurons and logistic maps as neurons in a simple odd-even structure of the input layer~\cite{as2023analysis}. HNL gave comparable performance to homogeneous NL and in the case of {\it Seeds} and {\it Haberman's Surival} datasets, it outperformed. We also studied the effect of degree of chaos on classification accuracies, as characterized by the lyapunov exponent of the chaotic 1D neurons in HNL. 

In this work, for the first time, we propose \verb|Random Heterogenous Neurochaos Learning| (RHNL) architecture. As noted in Table~\ref{table0:ann_bnn}), the human brain not only has heterogeneous neurons organized in layers, but there is an element of randomness involved. No two human brains have the same topological connectivity of neurons in their networks. The randomness is due to differences in early development that is a function of environment, learning and genetic factors. Inspired by this fact, we incorporate randomness and heterogeneity in NL. Specifically, we have analyzed three different RHNL architectures. The first one consists of $25 \%$ of logistic map neurons and the remaining $75 \%$ of GLS neurons, all randomly placed in the input layer. The second one consists of $50\% -50 \%$ of logistic-GLS neurons (again randomly placed) while the third architecture is composed of $75\% -25 \%$ of logistic-GLS neurons. We have rigorously tested these architectures (on classification tasks) in conjunction with different classifiers (cosine similarity and other ML classifiers) on a number of publicly available datasets. 


This paper is structured as follows. The proposed RHNL architecture is introduced for the first time (in Section 2). This is followed by a description of datasets in Section 3 and classifiers used in our study (Section 4). Experiments along with their results follow in Section 5. Section 5.1 gives the results obtained for Time Series Dataset. Discussion on the classification performance of $ChaosFEX_{RHNL}$ for debris and urban images are included in section 5.2. Results and analysis of the classification performance of $ChaosFEX_{RHNL}$ for brain tumor image dataset are included in section 5.3. Performance analysis of $ChaosFEX_{RHNL}$ in comparison with stand-alone ML classifiers in included in section 5.4. Section 5.5 contains the discussion on the performance of RHNL in low training sample regime. The paper then concludes with discussion, followed by potential research directions for the future in Section 6. The appendix contains the complete details of hyperparameter tuning of all the learning architectures used in this study.


\section{Proposed Architecture\label{sec_proposed_architecture}}
\begin{figure}[h]
			\centering
          		\includegraphics[scale=0.4]{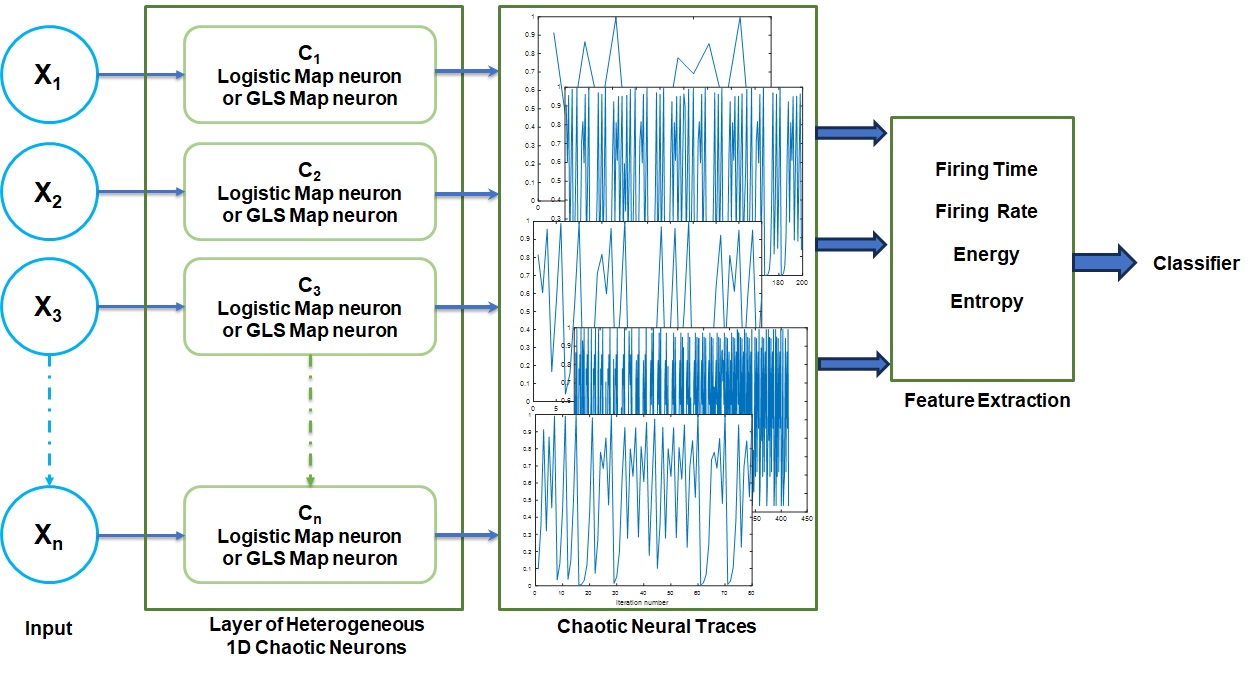}
  			\caption{\label{figure1:proposed architecture} Random Heterogenous Neurochaos Learning (RHNL) Architecture: ($X_{1}, X_{2}...X_{n}$) are the input stimuli (data sample), ($C_{1},C_{2},C_{3},....,C_{n-1},C_{n}$) are neurons which can  be either 1D Logistic map or GLS map. Each neuron fires chaotically until it detects the input stimuli. From the neural traces of every neuron (which is chaotic), four features namely firing-time, firing-rate, entropy and energy are extracted. These {\it ChaosFEX} features would now be fed to either a cosine similarity classifier or any of the standard machine learning classifiers. The logistic map and GLS neurons are randomly placed in the input layer with  one of the three following proportions: $25\%-75\%$, $50\%-50\%$ and $75\%-25\%$.}
  		\end{figure}

In order to mimic the randomness and heterogeneity of neuronal structures present in our brains, we propose a novel neurochaos learning architecture depicted in Figure~\ref{figure1:proposed architecture}. The input layer of this {\it Random Heterogeneous Neurochaos Learning} architecture (RHNL) consists of both chaotic 1D Logistic map and Generalized L\"{u}roth Series (GLS) map neurons, but at randomized locations. Contrast this with the {\it Heterogeneous Neurochaos Learning} or HNL proposed in ~\cite{remya2020analysis} where we had employed a simple odd-even structure (odd positions for GLS map neurons and even positions for logistic map neurons). In RHNL, we control the proportion of the randomly placed GLS and logistic neurons in the input layer to yield three distinct RHNL architectures: $25\%-75\%$, $50\%-50\%$ and $75\%-25\%$. In each case, the locations of the neurons are chosen uniformly at random. Following the flow in Figure~\ref{figure1:proposed architecture}, each neuron (either a GLS or a logistic map neuron) starts firing chaotically as soon as it encounters an input stimuli ($x_i$). Each input stimuli is a data (text/image/video etc.) sample of a particular class which RHNL is tasked to learn and classify. Each neuron stops firing as soon as it detects the input stimuli (when it lands in an $\epsilon$-neighbourhood). This completes a chaotic neural trace. Since different neurons detect their corresponding stimuli at different times, the chaotic neural traces are of unequal length (similar to the brain). These chaotic neural traces are then analyzed to extract features ({\it ChaosFEX}) such as entropy, energy, firing time and firing rate. Subsequently, the {\it ChaosFEX} features are fed to a classifier - that may consist of a basic {\it Cosine Similarity-based} classifier (see~\cite{balakrishnan2019chaosnet}) or one of the standard ML classifiers (see~\cite{sethi2023neurochaos}) such as SVM: Support Vector Machines, DT: Decision Trees, kNN: k-Nearest Neighbour, AB: AdaBoost, RF: Random Forests, or GNB: Gaussian Naive Bayes. 

Each neuron (logistic or GLS) starts of with a fixed {\it initial neural activity} value $q$ which is one of the hyperparameters of the learning algorithm. The other hyperparameters are the value of $\epsilon$ ({\it noise intensity}) that determines the stopping criteria of the neural firing/trajectory (which is chaotic) and the {\it Discrimination Threshold} ($b$) which is needed to compute Shannon Entropy (one of the {\it ChaosFEX} features) from the {\it symbolic sequence} of the chaotic neural trace~\cite{balakrishnan2019chaosnet, harikrishnan2020neurochaos, sethi2023neurochaos}. These three hyperparameters $(q,\epsilon,b)$ are determined by a cross-validation strategy (of five folds). We now describe the GLS and logistic map neurons.

\subsection{The Generalized L\"{u}roth Series  (GLS) Neuron}
In \cite{balakrishnan2019chaosnet}, the  one dimensional discrete dynamical system known as the GLS: Generalized L\"{u}roth Series is used as the neuron. Skew-tent/tent maps, skew-binary/binary maps are commonly used among the GLS maps. This class of one-dimensional systems/maps have demonstrated high effectiveness in various engineering applications~\cite{nagaraj4unreasonable}. In our proposed {\it Random Heterogenous Neurochaos Learning} (RHNL) architecture, skew-tent maps are used as chaotic neurons.  The Skew-tent map  $C_{\emph Skew-Tent} :[0.0,1.0) \mapsto [0.0,1.0) $ is mathematically defined as :
 \begin{equation}\label{skew tent map}
 \begin{aligned}
 C_{Skew-Tent}(z)=\left\{\begin{matrix}
\frac{z}{b} &,& 0\leq z< b,\\ 
 \frac{(1-z)}{(1-b)} &,&b\leq z< 1,
\end{matrix}\right.
 \end{aligned}
\end{equation}\\
where $z$ $\epsilon$ $[0.0, 1.0)$ and $0.0<b<1.0$. 
  \subsection{The Logistic Map Neuron}	
\begin{figure}[!h]
			\centering
    		\includegraphics[scale=0.14]{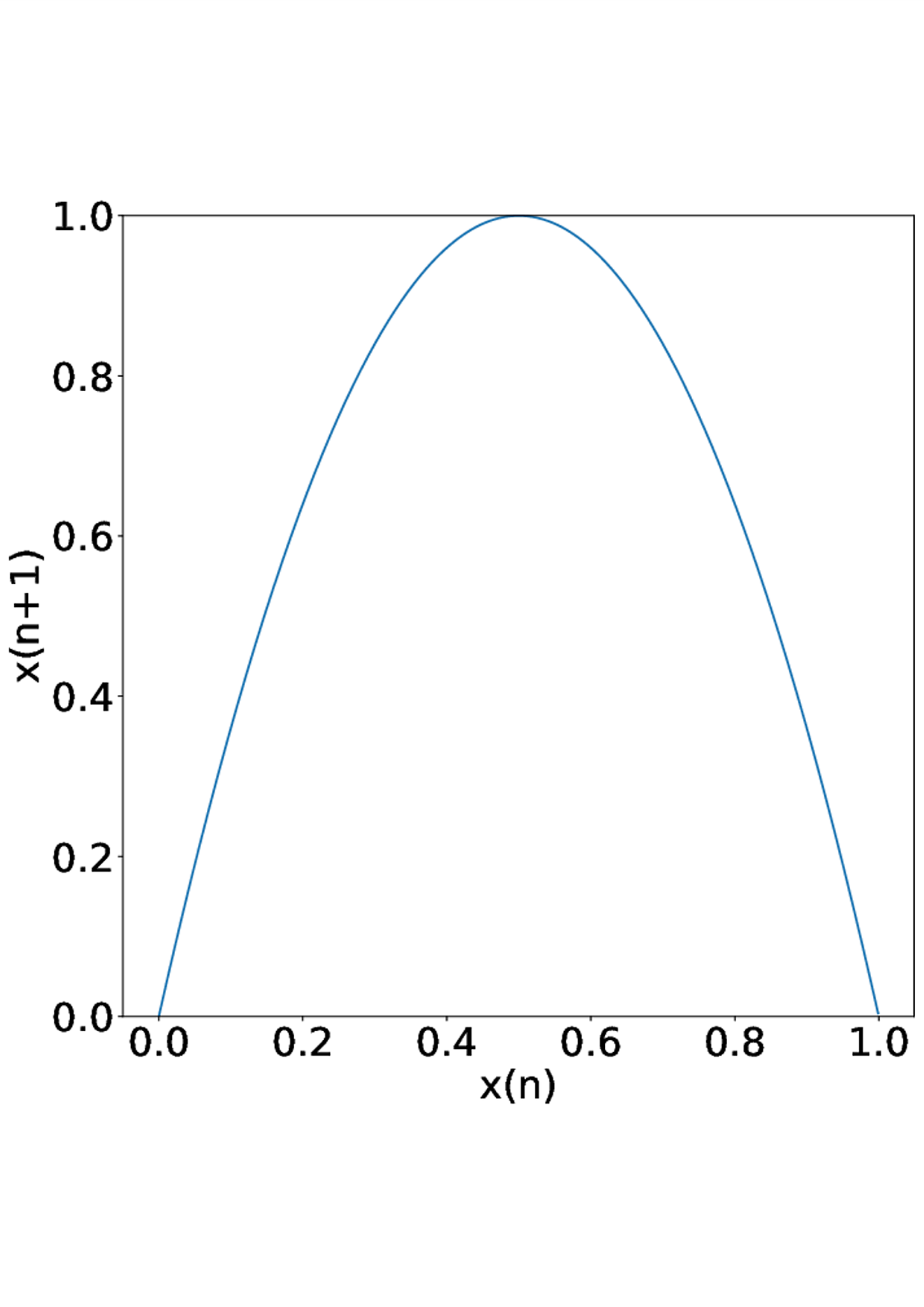}
    		\hspace{0.2in}
    		 \includegraphics[scale=0.14]{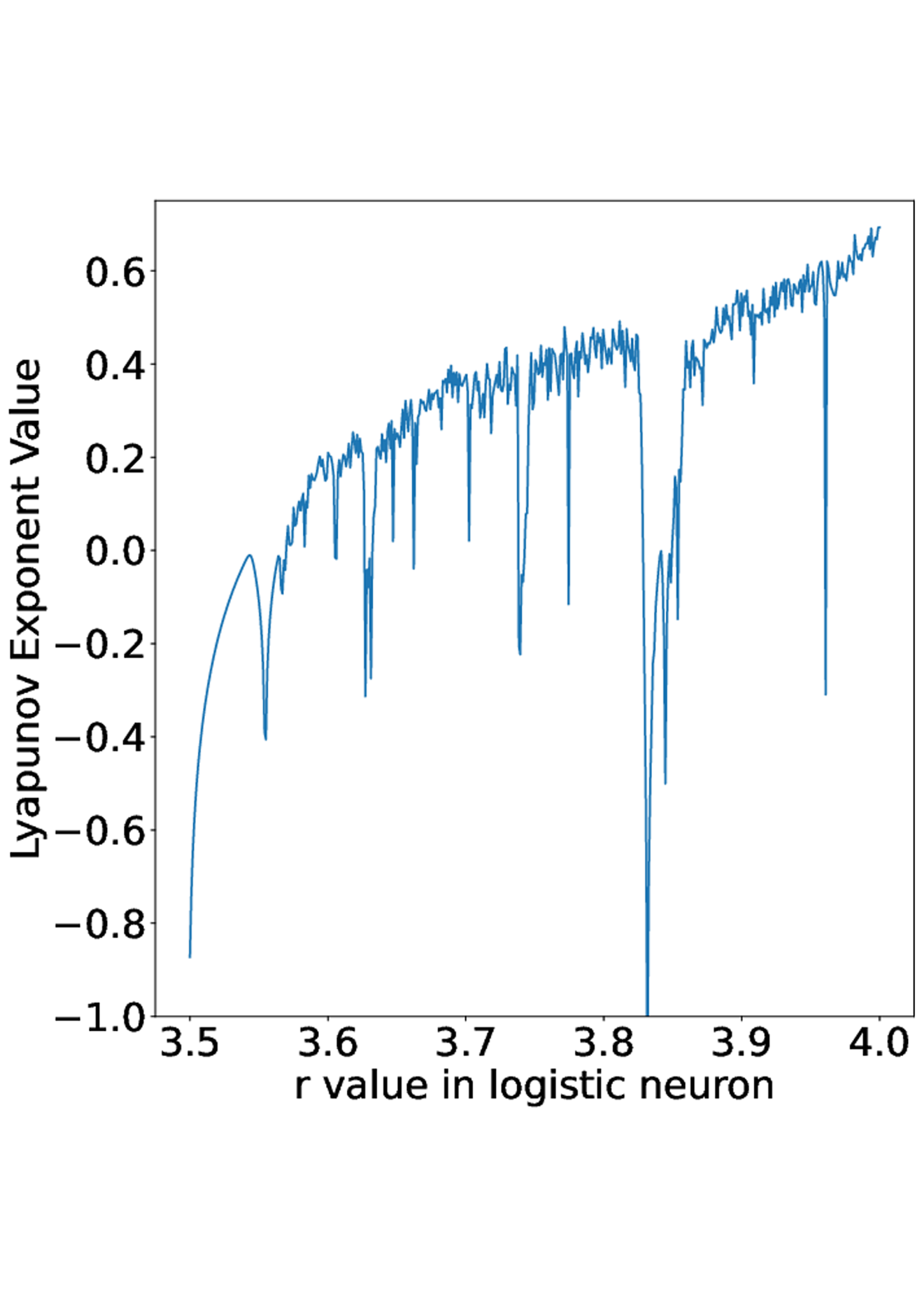}
  			\caption{\label{figure2:logisticmapfunction} (I) Left: One dimensional logistic map  with $r$ value set to $4.0$. (II) Right: The lyapunov exponent computed for different $r$ values (varied from $3.5$ to $4.0$). The initial neural activity $q$ was set to $0.01$.}
  		\end{figure}

The one-dimensional dynamical system/map known as the {\it Logistic map} is arguably the simplest example of a chaotic map~\cite{phatak1995logistic}. We explore the use of this one-dimensional map/dynamical system (in chaotic state) as neurons for RHNL. The equation for this dynamical system/map is:
\begin{equation}
x_{z+1}=rx_{z}(1-x_{z}),
\label{logisticmapequation}
\end{equation}
where $0.0 \leq x_z < 1.0$ and the bifurcation parameter is: $0 < r \leq 4.0$. $z$ is the iteration/time step. It is widely recognized that the logistic map displays chaotic behavior for $r$ values that are greater than $3.56995$. However, there exist certain regions of $r$ referred to as {\it islands of stability} where chaotic behavior is lost. Figure~\ref{figure2:logisticmapfunction} shows the first return map of the logistic dynamical system/map with $r$ set to $4.0$. The lyapunov exponent is also plotted alongside.

Logistic map displays an infinite number of periodic orbits for each integer value of the period (only for specific values of $r$), indicating intricate dynamics~\cite{alligood1998chaos}. For numerous $r$ values, this map exhibits high sensitivity to initial conditions ({\it Butterfly Effect}) -- a characteristic hallmark/feature of chaos. The degree of sensitivity to initial values can be quantified through the {\it Lyapunov exponent}. A  lyapunov exponent value greater than zero is a symptom of chaotic behavior. Alternatively, a lyapunov exponent $=0$  or $<0$ suggests either a periodic/eventually periodic/quasi-periodic behaviour. For the difference equation (first-order):
\begin{equation}
x_{j+1}=G(x_j),
\end{equation}
the lyapunov exponent is defined as: 
\begin{equation}
{\lambda_G(x)}= \lim_{k\to\infty} {\frac {1}{k}} \sum_{j=0}^{k-1}\ln |G^{'}(x_j)|,
\end{equation}
where $G(\cdot)$ is assumed to be differentiable. The initial value $x_0$ is randomly chosen (from an uniform distribution) to lie between $0.0$ and $1.0$. $x_0 \rightarrow G(x_0) \rightarrow G^2(x_0) \rightarrow \ldots$ is the trajectory.  $G(x)$ is given by equation~\ref{logisticmapequation} for the logistic map/dynamical system.


\subsection{Feature Transformation, Extraction and Classification}
As depicted in Figure~\ref{figure1:proposed architecture}, in the newly proposed RHNL structure, both 1D chaotic Logistic and GLS neurons are employed in randomly selected locations of input layer. Each neurons transforms the input stimulus (data sample) into respective chaotic neural traces (or chaotic trajectories). From these trajectories, required features are derived for further classification process. The quantity of neurons, denoted as ($C_{1}, C_{2}, ....C_{n}$) as illustrated in Figure~\ref{figure1:proposed architecture}, will precisely match the quantity of features of the input data/samples. All input layer neurons independently start firing (when an input value/stimulus triggers it), say $x_i$. These start of with an initial value/neural-activity of $q$ units. Each of the input values are scaled to fall within the range of $[0,1]$.  Upon entering the open $\epsilon$-ball (or neighbourhood) of the input stimulus, the neural trace comes to a halt. In the context of classification using Neurochaos Learning (NL) architecture, a straightforward decision rule is employed, relying on {\it mean-representation vectors} (see~\cite{balakrishnan2019chaosnet}). Tuning of the three essential hyper-parameters is necessary: noise intensity ($\epsilon$), initial neural-activity ($q$) and discrimination-threshold ($b$). To determine the optimal performance, a cross-validation approach (with five-folds) is utilized for hyperparameter tuning. Following the tuning and stabilization of these for a specific data-set, we generate the {\it neurochaos} features. These features that are obtained from the neural trajectories/traces (which are chaotic) of the input layer neurons of the NL architecture (could be homogeneous NL or HNL or RHNL) are referred to as {\it ChaosFEX}~\cite{sethi2023neurochaos} features. These include firing-rate, firing-time, shannon entropy and energy (briefly described below).

Firing time is defined as the duration of the chaotic neural trajectory/trace to align with input value (or stimulus)~\cite{sethi2023neurochaos}. This duration is measured in terms of iteration steps. The firing rate is determined by that fraction of time during which the chaotic neural trajectory is greater than the discrimination threshold~\cite{sethi2023neurochaos}.
Energy of the chaotic trace/trajectory $c(t)$ is defined as:
\begin{equation}
{E_{c}}= \sum_{t=1}^{M} |c(t)|^2, 
\end{equation}
where $M =$ firing-time. Let symbolic sequence (binary) of the trace/trajectory, be denoted by   $s(t)$, which is expressed as:
\begin{equation}
s(t_{j})=\left\{\begin{matrix}
0,&&c(t_{j})< b \\ 
1,&&b\leqslant c(t_{j})<1, 
\end{matrix}\right.
\end{equation}
where $j=1$ to $M$ (the firing-time). The shannon first-order entropy for $s(t)$ is determined using the following calculations:
\begin{equation}
{H(s)}= - \sum_{t=1}^{2} p_i \log_2(p_i)~ bits,
\end{equation}
where $p_1$ corresponds to the probability of symbol $0$ while $p_2$ pertains to the probability of symbol $1$. Dimensions of these features are similar to the size of the inputs. For an input of dimensions $m \times n$, the output features will have a size of $m \times 4n$. This process involves converting the input data into a feature space characterized by high-dimensional chaos.

In the training phase of the algorithm, for each data sample of each class, a representation vector is formed with the set of {\it ChaosFEX} features namely  energy, entropy, firing time and firing rate. These representation vectors of all the data samples within a particular class are collected  and its mean is computed to yield a single {\it mean representation vector} for each class. Thus every class has its own distinct mean representation vector that is like a signature of the neurochaos features of that class. These mean representation vectors are fed to a classifier to perform the classification. In the testing phase, when a test sample appears at the NL architecture's input layer, it undergoes a similar transformation to yield the {\it ChaosFEX} features. These features are compared with the mean representation vectors of each class to determine the {\it closest} one and the class label of the closet one is declared to be the class of input test sample. For the \verb|ChaosNet| version of NL, we employ the {cosine similarity} measure to determine closeness. Alternatively, we could pass on the representation vectors corresponding to the data samples (within a class) to a traditional ML classifier and learn the decision boundary. 

 
\section{Datasets}
\label{section:Dataset}
To conduct our analysis, we picked datasets spanning various application domains. Before feeding the data samples to the chaotic neurons of  RHNL's input layer, the data samples of the input is normalized to lie within $[0,1]$. Class labels are assigned numerical names, starting from $0$. Further information about the datasets utilized can be found in Table~\ref{table1:dataset_details}. The data-sets are described briefly here.

\begin{table}[h!]
\caption{Datasets (with their details) employed in our study. \label{table1:dataset_details}}
\resizebox{\textwidth}{!}{%
\begin{tabular}{ccccc}
\hline
\textbf{Data-set}        & \textbf{Num. of Classes} & \textbf{Samples per class (Training)} & \textbf{Samples per class (Testing)} & \textbf{Ref.} \\ \hline
Iris                    & 3                          & (40, 41, 39)                      & (10, 9, 11)                      & \cite{fisher1936use,dua2017uci}            \\ \hline
Ionoshpere              & 2                          & (98, 182)                        & (28, 43)                        & \cite{sigillito1989classification,dua2017uci}             \\ \hline
Wine                    & 3                          & (45, 57, 40)                      & (14, 14, 8)                      & \cite{vandeginste1990parvus,dua2017uci}             \\ \hline
Bank Note Authentication & 2                          & (614, 483)                       & (148, 127)                      & \cite{gillich2010banknote,dua2017uci}            \\ \hline
Haberman's Survival     & 2                          & (181, 63)                        & (44, 18)                        & \cite{haberman1973analysis,dua2017uci}           \\ \hline
Breast Cancer Wisconsin & 2                          & (367, 193)                       & (91, 48)                        & \cite{street1993nuclear,dua2017uci}           \\ \hline
Statlog(Heart)          & 2                          & (117, 99)                        & (33, 21)                        &  \cite{dua2017uci}                \\ \hline
Seeds                   & 3                          & (59, 56, 53)                      & (59, 56, 53)                     & \cite{dua2017uci}             \\ \hline
FSDD&10&(40, 35, 44, 42, 
38, 34, 37, 44, 33, 37)&(10, 15, 6, 8, 8, 7, 13, 6, 10, 13)&\cite{jackson2018jakobovski}\\ \hline
\end{tabular}%
}
\end{table}

The {\it Iris} dataset~\cite{fisher1936use,dua2017uci} comprises $150$ instances distributed across $3$ classes: {\it Setosa, Versicolour}, and {\it Virginica}. The classification features include sepal length, sepal width,  petal length, and petal width. The distribution of data instances into train and test sets for our analysis is presented in Table~\ref{table1:dataset_details}. The {\it Ionosphere} dataset~\cite{sigillito1989classification,dua2017uci} is divided into $2$ classes, labeled as $Good$ or $Bad$. Radar signal reflects back if any structure is present in the ionosphere. This state is represented as `$Good$'. `$Bad$' denotes the condition in which the ionosphere is penetrated by this (radar) signal. The data-set comprises $126$ instances labeled as $Good$, $225$ instances labeled as $Bad$, and includes $34$ attributes. The distribution of instances into train and test sets for our analysis is outlined in Table ~\ref{table1:dataset_details}. The {\it Wine} dataset~\cite{vandeginste1990parvus,dua2017uci} comprises $178$ instances categorized into $3$ classes labeled as $1, 2$, and $3$. The chemical constituents of each data are considered for classification. The distribution of instances into train and test sets for our analysis is presented in Table ~\ref{table1:dataset_details}.


The {\it Bank Note Authentication} dataset~\cite{gillich2010banknote,dua2017uci}  has two classes:$Genuine$ or $Forgery$ based on the images of the banknotes. Wavelet transformation is applied on images and the features such as variance, skewness, curtosis, and entropy are derived. In total, the dataset comprises $1372$ instances, with $762$ instances classified as $Genuine$ and $610$ instances as $Forgery$. The distribution of instances into train and test sets for our analysis is outlined in Table ~\ref{table1:dataset_details}. The {\it Haberman’s Survival} dataset~\cite{haberman1973analysis,dua2017uci} encompasses three attributes (collected from those patients who underwent breast cancer surgery). The label$1$ class represent the patients survived for $\geq 5$ years. Class $2$ represent the patients who died in a span of $5$ years. The train and test sets distribution for our analysis is presented in Table ~\ref{table1:dataset_details}. Nine parameters are considered for {\it Breast Cancer Wisconsin} dataset~\cite{street1993nuclear,dua2017uci}. Data instances are categorized to be either $Malignant$ or $Benign$. In total, there are 699 instances, with 241 being classified as $Malignant$ and 458 as $Benign$. The distribution of instances into train and test sets for our analysis is outlined in Table ~\ref{table1:dataset_details}.


The {\it Statlog (Heart)} dataset~\cite{dua2017uci} contains two classes of data: patients having heart problems are classified in $Class$-$1$ and patients without any heart disease are represented in $Class$-$2$. The distribution of instances into train and test sets for our analysis is presented in Table ~\ref{table1:dataset_details}. The {\it Seeds} dataset~\cite{dua2017uci} is employed to distinguish between three types of wheat, namely $Kama$, $Rosa$, and $Canadian$. Wheat kernels are used for identification among the types with seven parameters that represent various features of the kernels. In all, 210 data instances are in consideration, with 70 instances allocated for each class. The distribution of instances into train and test sets for our analysis is outlined in Table ~\ref{table1:dataset_details}.


In our analysis, we also included a time series dataset namely {\it Free Spoken Digit Dataset (FSDD)}. Recordings of six speakers reciting numbers from $0$ to $9$~\cite{jackson2018jakobovski} is contained in this data-set. There are $50$ recordings for each number per speaker. The samples in the {\it FSSD } data set are preprocessed using fast fourier transform (or FFT). We considered {\it Jackson} (one of the speakers) that has instances of data numbering $500$. We filtered  $480$ data instances  for feeding into our proposed algorithms and analysed. 
For our analysis, the train/test sets of {\it FSDD} are shown in
Table~\ref{table1:dataset_details}.

Around $85$ images of debris scars and urban settlements from five Asian countries (India, Nepal, Japan, Taiwan and China) were obtained from Planet labs~\cite{Planet} imagery with $3-5$m resolution for our analysis to identify the classification performance of RHNL, specifically  ChaosFEX$_{RH25L75G}$ architecture. For our analysis, we also have considered $100$ MRI brain images from {\it Kaggle} online repository~\cite{Kaggle}.

\section{Classifiers\label{details of classifiers}}
As mentioned previously, RHNL supports the use of traditional popular ML classifiers. The neurochaos features ({\it ChaosFEX}) can be fed to one of the many widely available machine learning classifiers to perform classification. Previously, it has been demonstrated that neurochaos features boost the performance of the ML classifiers (see~\cite{sethi2023neurochaos}). In our study we use the following ML classifiers:  Support Vector Machine (SVM)~\cite{boser1992training}, AdaBoost~\cite{schapire2013explaining}, Decision Tree~\cite{quinlan1986induction}, Guassian Naive Bayes~\cite{berrar2018bayes}, k-NN~\cite{cover1967nearest} and Random Forests(RF)~\cite{breiman2001random}. Whenever a traditional ML classifier was used on the neurochaos featuers, the hyperparameters ($q$, $b$, $ \epsilon$) that were already tuned for the various RHNL architectures ($ChaosFEX$) were maintained and only the ML hyperparameters are further tuned. This reduces the computational burden.

The Adaptive Boosting (AdaBoost) classifier has the following hyperparameters:
\begin{itemize}
\item \textbf{$n\_estimator$:} The maximum limit on the number of estimators at which boosting is terminated.
\end{itemize}
All other hyperparameters are maintained at their default values provided by {\it scikit-learn}. Tuned hyperparameters for all the datasets for RHNL that uses the AdaBoost classifier are given in Tables~\ref{table:25L75G:AdaBoost Hyperparameters},~\ref{table:50L50G AdaBoost Hyperparameters} and~\ref{table:75L25G:AdaBoost}. 

The Decision Tree (DT) classifier has the following hyperparameters: $min\_samples\_leaf$ (varied from $1$ to $10$ in increments of $1$), $max\_depth$ ($1$ to $10$ in increments of $1$), and $ccp\_alpha$.

All other hyperparameters are maintained at their default values provided by {\it scikit-learn}. The tuned hyperparameters for all the datasets for RHNL that uses the Decision Trees classifier are given in Tables ~\ref{table:25L75G:Decision Tree},~\ref{table:50L50G:Decision Tree}and ~\ref{table:75L25G DecisionTree}. 

The k-Nearest Neighbours (k-NN) classifier has the value of $k$ as a hyperparameter. This is varied from $1.0$ to $6.0$ (in incremets of $2$). All other hyper-parameters are maintained at their default values provided by {\it scikit-learn}. The tuned hyperparameters for all the datasets for RHNL that uses the kNN classifier are given in Tables~\ref{table:25L75G:knn },~\ref{table:50L50G:knn } and\ref{table:75L25G:knn }. 

The Random Forests (RF) classifier has the following hyperparameters: n\_estimators (can take values in the set  $\{1, 10, 100, 1000, 10000\}$), max\_depth ($1$ to $10$ in steps of $1$). All other hyperparameters are maintained at their default values provided by {\it scikit-learn}. The tuned hyperparameters for all the datasets for RHNL that uses the Random Forests classifier are given in Tables~\ref{table:25L75G:Random Forests}, \ref{table:50L50G:Random Forests} and \ref{table:75L25G:Random Forests}. 

For Support Vector Machines classifier (SVM), all hyperparameters (offered for linear support vector classification) are maintained at their default values provided by {\it scikit-learn}.

Gaussian Naive Bayes (GNB) classifier calculates the likelihood and prior probabilities for making predictions. GNB assumes that the features follow a Gaussian distribution. Default parameters offered for GNB by scikit-learn are retained for our analysis.

For ease of understanding, Table~\ref{table:RHNL_flavors} gives a summarized view of all the different neurochaos learning (NL) architectures with the corresponding notations that are used in this paper.

\begin{table}[h!]
\centering
\caption{Various learning architectures of Neurochaos Learning (NL). These include homogenous NL, heterogeneous NL (HNL) and random heterogeneous NL (RHNL), including combinations with ML classifiers.\label{table:RHNL_flavors}}
\begin{tabular}{p{0.5cm} p{1.8cm} p{3.0cm} p{3.5cm} p{2.5cm} p{0.5cm}}
\hline
\textbf{No.} & \textbf{NL Architecture} & \textbf{Type of Neurons}  & \textbf{Notation}  &  \textbf{Classifiers} & \textbf{Ref.} \\
\hline
1 & ChaosNet & Homogeneous, GLS & ChaosFEX & Cosine similarity & \cite{balakrishnan2019chaosnet} \\
\hline
2 & ChaosNet & Homogeneous, Logistic & ChaosFEX$_{logistic}$ & Cosine similarity & \cite{remya2020analysis} \\
\hline
3 & NL & Homogeneous, GLS & ChaosFEX$+$ML & SVM, AB, DT, kNN, GNB, RF & \cite{sethi2023neurochaos} \\
\hline
4 & NL & Homogeneous, Logistic & ChaosFEX$_{logistic}+$ML & SVM, AB, DT, kNN, GNB, RF & \cite{remya2020analysis} \\
\hline
5 & HNL: ChaosNet & Heterogeneous, GLS, Logistic in odd-even structure & ChaosFEX$_{Hetero}$ & Cosine similarity & \cite{remya2020analysis} \\
\hline
6 & HNL & Heterogeneous, GLS, Logistic in odd-even structure  & ChaosFEX$_{Hetero}+$ML & SVM, AB, DT, kNN, GNB, RF & \cite{remya2020analysis} \\
\hline
7 & RHNL: ChaosNet & Heterogeneous \& Random, GLS, Logistic in randomized locations & ChaosFEX$_{RH25L75G}$, ChaosFEX$_{RH50L50G}$, ChaosFEX$_{RH75L50G}$  & Cosine similarity & This work. \\
\hline
8 & RHNL & Heterogeneous \& Random, GLS, Logistic in randomized locations & ChaosFEX$_{RH25L75G}+$ML, ChaosFEX$_{RH50L50G}+$ML, ChaosFEX$_{RH75L50G}+$ML  & SVM, AB, DT, kNN, GNB, RF & This work. \\
\hline

\end{tabular}%
\end{table}

For RHNL, as noted in Table~\ref{table:RHNL_flavors}, ChaosFEX$_{RH25L50G}$, ChaosFEX$_{RH50L50G}$, and ChaosFEX$_{RH75L25G}$ refers to the three distinct random heterogeneous neurochaos learning architectures with $25\%-75\%$, $50\%-50\%$ and $75\%-25\%$ proportion of 1D chaotic Logistic neurons and GLS neurons respectively. These chaotic neurons are placed at random locations in the input layer of RHNL. 

\section{Experiments and Results}
\label{section:Experiments and Results}
The performance of ChaosFEX$_{RH}$ architectures are analysed for various datasets using Macro F1-score (a function of both macro Recall as well as macro Precision). True-Positive rate $(TP)$ signifies a positive target-value correctly identified as Positive. True-Negative rate $(TN)$ denotes a `negative target-value' correctly classified as Negative. False Positive rate $(FP)$ accounts those instances when a `negative target value' is inaccurately deemed/classified as Positive. False-Negative rate $(FN)$ accounts those instances when a `positive target value' is erroneously deemed/classified as Negative. Mathematically, they are described as:

\begin{equation}
    Accuracy = \frac{(TP+TN)}{(TP+TN+FP+FN)},
    \end{equation}
    
    \begin{equation}
Precision = \frac{TP}{(TP+FP)},
\end{equation}
   \begin{equation}
Recall = \frac{TP}{(TP+FN)},
\end{equation}
   \begin{equation}
F1 = 2.0 \times \frac{Precision \times Recall}{Precision+Recall}.
\end{equation}

The Macro F1-score is computed as the average of all F1-scores (for the $m$ classes), given by:
\begin{equation}
    Macro \ F1-score = \frac{F1 Class_1+F1 Class_2+\ldots+F1 Class_m}{m}.
    \end{equation}

Tuning of the 3 hyper-parameters ($q, b, \epsilon$) are performed across various datasets for the three RHNL architectures proposed namely ChaosFEX$_{RH25L75G}$, ChaosFEX$_{RH50L50G}$  and ChaosFEX$_{RH75L25G}$. Tuned values of hyperparameters for all architectures are given in tables~\ref{table:25L75G:Hyperparameters}, ~\ref{table:50L50G:Hyperparamters} and ~\ref{table:75L25G:Hyperparamters}. Cross validation using five folds is utilized to fine tune hyper-parameters and determine the best achieved performance.  

We also analysed the performance of our proposed RHNL architectures in combinations with other traditional ML classifiers such as AdaBoost, Decision Trees, Gaussian Naive Bayes, kNN  and Random Forests. The ML classifier parameters are tuned for various datasets to obtain the best accuracy possible.

\begin{table}[h!]
\centering
\caption{Tuned hyperparameters for ChaosFEX$_{RH25L75G}$ for the eight datasets. 
\label{table:25L75G:Hyperparameters}}
\begin{tabular}{@{}lccc@{}}
\toprule
\textbf{Data-set}                & \textbf{\begin{tabular}[c]{@{}c@{}}$q$\end{tabular}} & \textbf{\begin{tabular}[c]{@{}c@{}}$b$\end{tabular}} & \textbf{\begin{tabular}[c]{@{}c@{}}$\epsilon$\end{tabular}} \\ \midrule
Iris & 0.062 & 0.185& 0.298  \\ \midrule
Ionosphere & 0.010 & 0.409 & 0.051  \\ \midrule
Wine & 0.460 & 0.469 & 0.141 \\ \midrule
Bank-Note-Authentication  & 0.360  & 0.419  & 0.121      \\ \midrule
Haberman's-Survival & 0.050  & 0.269 & 0.031          \\ \midrule
Breast-Cancer-Wisconsin  & 0.170  & 0.460 & 0.050    \\ \midrule
Statlog (Heart)  & 0.470 & 0.489 & 0.030   \\ \midrule
Seeds & 0.050 & 0.189 & 0.161    \\ \bottomrule
\end{tabular}
\end{table}
\begin{table}[h!]
\centering
\caption{Tuned hyperparameters for ChaosFEX$_{RH50L50G}$ for the eight datasets. 
\label{table:50L50G:Hyperparamters}}
\begin{tabular}{@{}lccc@{}}
\toprule
\textbf{Data-set}                & \textbf{\begin{tabular}[c]{@{}c@{}}$q$\end{tabular}} & \textbf{\begin{tabular}[c]{@{}c@{}}$b$\end{tabular}} & \textbf{\begin{tabular}[c]{@{}c@{}}$\epsilon$\end{tabular}} \\ \midrule
Iris & 0.050 & 0.359  & 0.221  \\ \midrule
Ionosphere & 0.099  & 0.479   & 0.061   \\ \midrule
Wine        & 0.460  & 0.469  & 0.131  \\ \midrule
Bank-Note-Authentication & 0.090  & 0.289 & 0.041  \\ \midrule
Haberman's-Survival & 0.140 & 0.489 & 0.021  \\ \midrule
Breast-Cancer-Wisconsin & 0.069  & 0.139 & 0.041  \\ \midrule
Statlog (Heart) & 0.180  & 0.169 & 0.011   \\ \midrule
Seeds & 0.050 & 0.139 & 0.151 \\ \bottomrule
\end{tabular}
\end{table}
\begin{table}[h!]
\centering
\caption{Tuned hyperparameters for ChaosFEX$_{RH75L25G}$ for the eight datasets. 
\label{table:75L25G:Hyperparamters}}
\begin{tabular}{@{}lccc@{}}
\toprule
\textbf{Data-set}                & \textbf{\begin{tabular}[c]{@{}c@{}}$q$\end{tabular}} & \textbf{\begin{tabular}[c]{@{}c@{}}$b$\end{tabular}} & \textbf{\begin{tabular}[c]{@{}c@{}}$\epsilon$\end{tabular}} \\ \midrule
Iris & 0.15   & 0.299  & 0.231   \\ \midrule
Ionosphere  & 0.02 & 0.219 & 0.809  \\ \midrule
Wine & 0.47& 0.479 & 0.131  \\ \midrule
Bank-Note-Authentication & 0.01 & 0.259 & 0.071  \\ \midrule
Haberman's-Survival & 0.23 & 0.1 & 0.011     \\ \midrule
Breast-Cancer-Wisconsin & 0.14 & 0.489  & 0.021   \\ \midrule
Statlog (Heart) & 0.13 & 0.1 & 0.051   \\ \midrule
Seeds  & 0.05 & 0.189 & 0.151\\ \bottomrule
\end{tabular}
\end{table}

Macro F1 scores obtained for various datasets with ChaosFEX$_{RH25L75G}$ and ChaosFEX$_{RH25L75G}$+SVM are reported in Table~\ref{table:25L75G:F1}. For {\it Haberman's Survival} dataset, we achieved an improved macro F1 score of $0.73$ compared to the best F1 score reported in earlier works~\cite{sethi2023neurochaos,as2023analysis}. Macro F1 score of {\it Statlog (Heart)} dataset is also increased to $0.84$ with ChaosFEX$_{RH25L75G}+$SVM classifier. 

Table~\ref{table:50L50G:F1} gives the macro F1 score obtained with ChaosFEX$_{RH50L50G}$ and ChaosFEX$_{RH50L50G}+$SVM. Table~\ref{table:75L25G:F1} gives the macro F1 score obtained with ChaosFEX$_{RH75L25G}$ and ChaosFEX$_{RH75L25G}+$SVM. 

\begin{table}[h!]
\centering
\caption{Macro F1 scores reported for ChaosFEX$_{RH25L75G}$ and ChaosFEX$_{RH25L75G}+$SVM.
\label{table:25L75G:F1}}
\begin{tabular}{@{}ccc@{}}
\toprule
\textbf{Data-set}                         & \textbf{$ChaosFEX_{RH25L75G}$} & \textbf{$ChaosFEX_{RH25L75G}$+SVM} \\ \midrule
Iris  & 1  & 1                             \\ \midrule
Ionosphere  & 0.6  & 0.88   \\ \midrule
Wine  & 0.6 & 0.94 \\ \midrule
Bank-Note-Authentication & 0.75 & 0.9  \\ \midrule
Haberman's-Survival & 0.73 & 0.56 \\ \midrule
Breast-Cancer-Wisconsin  & 0.85  & 0.98  \\ \midrule
Statlog(Heart) & 0.77 & 0.84    \\ \midrule
Seeds  & 0.81    & 0.84    \\ \bottomrule
\end{tabular}
\end{table}

\begin{table}[h!]
\centering
\caption{Macro F1 scores reported for ChaosFEX$_{RH50L50G}$ and ChaosFEX$_{RH50L50G}+$SVM.
\label{table:50L50G:F1}}
\begin{tabular}{@{}ccc@{}}
\toprule
\textbf{Data-set Name}                         & \textbf{$ChaosFEX_{RH50L50G}$} & \textbf{$ChaosFEX_{RH50L50G}$+SVM} \\ \midrule
Iris                                          & 1                         & 1                             \\ \midrule
Ionosphere                                    & 0.58                      & 0.9                           \\ \midrule
Wine                                          & 0.59                      & 0.94                          \\ \midrule
Bank-Note-Authentication                      & 0.59                      & 0.72                          \\ \midrule
Haberman's-Survival                           & 0.68                      & 0.47                          \\ \midrule
Breast-Cancer-Wisconsin                       & 0.77                      & 0.92                          \\ \midrule
Statlog(Heart)                                & 0.78                      & 0.79                          \\ \midrule
Seeds                                          & 0.72                      & 0.81                          \\ \bottomrule
\end{tabular}
\end{table}

\begin{table}[h!]
\centering
\caption{Macro F1 scores reported for ChaosFEX$_{RH75L25G}$ and ChaosFEX$_{RH75L25G}+$SVM.
\label{table:75L25G:F1}}

\begin{tabular}{@{}lccc@{}}
\textbf{Data-set Name}                         & \textbf{$ChaosFEX_{RH75L25G}$} & \textbf{$ChaosFEX_{RH75L25G}$+SVM} \\ \midrule
Iris & 1  & 0.97 \\ \midrule
Ionosphere & 0.71   & 0.94   \\ \midrule
Wine  & 0.63     & 0.97    \\ \midrule                      
Bank-Note-Authentication  & 0.65   & 0.84 \\ \midrule
Haberman's-Survival  & 0.6 & 0.51   \\ \midrule
Breast-Cancer-Wisconsin  & 0.79 & 0.94  \\ \midrule
Statlog(Heart)  & 0.65  & 0.85  \\ \midrule
Seeds  & 0.78  & 0.86    \\ \bottomrule                    
\end{tabular}
\end{table}

\newpage
Random Heterogenous Neurochaos Learning architectures which incorporate ChaosFEX features with other ML classifiers such as AdaBoost (AB), Decision Trees (DT), k-NN, Gaussian Naive Bayes (GNB), and Random Forests (RF) are implemented and the results indicate that {\it randomness} and {\it heterogeneity} introduced in the NL architectures  yields superior performance when compared with homogeneous or fixed heterogeneous structures.

The macro F1 scores obtained for $ChaosFEX_{RH25L75G}$+AdaBoost, $ChaosFEX_{RH50L50G}$+AdaBoost and $ChaosFEX_{RH75L25G}$+AdaBoost structures are given in Table~\ref{table:Macro F1 Score AdaBoost}. Accuracy of 100\% is obtained for {\it Wine} dataset with $ChaosFEX_{RH50L50G}$+AdaBoost architecture. Macro F1-score $=0.99$ is successfully achieved for {\it Bank Note authentication} data-set for both $ChaosFEX_{RH50L50G}$+AdaBoost and $ChaosFEX_{RH75G25L}$+AdaBoost. High F1-score $=0.99$ is also achieved for {\it Breast Cancer Wisconsin} data-set when $ChaosFEX_{RH75L75G}$+AdaBoost is implemented.


\begin{table}[h!]
\centering
\caption{Macro F1 scores obtained for different ChaosFEX$_{RH}+$AdaBoost architectures.\\ {\bf Bold} fonts indicate the highest F1 score achieved for the respective dataset.}
\label{table:Macro F1 Score AdaBoost}
\begin{tabular}{llll}
\hline
\textbf{Data-set Name} & $ChaosFEX_{RH25L75G}$+AB & $ChaosFEX_{RH50L50G}$+AB& $ChaosFEX_{RH75L25G}$+AB \\ \hline
Iris & \textbf{1} & \textbf{1} & 0.967 \\ \hline
Ionosphere & \textbf{0.97} & \textbf{0.97} & \textbf{0.97} \\ \hline
Wine & 0.97 & \textbf{1} & 0.944 \\ \hline
Bank-Note-Authentication & 0.93 & \textbf{0.99} & 0.989 \\ \hline
Haberman's-Survival & 0.5 & 0.56 & \textbf{0.66} \\ \hline
Breast-Cancer-Wisconsin & 0.98 & 0.98 & \textbf{0.99} \\ \hline
Statlog(Heart) & 0.81 & 0.85 & \textbf{0.88} \\ \hline
Seeds & \textbf{0.86} & 0.77 & 0.73 \\ \hline
\end{tabular}%
\end{table}


The macro F1 scores obtained for $ChaosFEX_{RH25L75G}+$Decision Trees, $ChaosFEX_{RH50L50G}+$Decision Trees and $ChaosFEX_{RH75L25G}+$Decision Trees can be found in Table~\ref{table:Macro F1 Score Decision Tree}.  High F1 score $=0.98$ for {\it Breast Cancer Wisconsin} data-set with $ChaosFEX_{RH25L75G}+$Decision Trees and  $ChaosFEX_{RH75L25G}+$Decision Trees has been achieved.
%
\begin{table}[h!]
\centering
\caption{Macro F1 score obtained for different ChaosFEX$_{RH}+$Decision Trees architectures.\\{\bf Bold} fonts indicate the highest F1 score achieved for the respective dataset.}
\label{table:Macro F1 Score Decision Tree}
\begin{tabular}{llll}
\hline
\textbf{Data-set Name} & $ChaosFEX_{RH25L75G}$+DT & $ChaosFEX_{RH50L50G}$+DT & $ChaosFEX_{RH75L25G}$+DT \\ \hline
Iris & \textbf{1} & 0.97 & 0.97 \\ \hline
Ionosphere & 0.92 & 0.91 & \textbf{0.97} \\ \hline
Wine & \textbf{0.95} & 0.94 & \textbf{0.95} \\ \hline
Bank-Note-Authentication & \textbf{0.95} & 0.90 & 0.89 \\ \hline
Haberman's-Survival & 0.60 & \textbf{0.65} & 0.63 \\ \hline
Breast-Cancer-Wisconsin & \textbf{0.98} & 0.97 & \textbf{0.98} \\ \hline
Statlog(Heart) & \textbf{0.92} & 0.84 & 0.86 \\ \hline
Seeds & \textbf{0.81} & \textbf{0.81} & 0.76 \\ \hline
\end{tabular}%
\end{table}


The macro F1 scores obtained for $ChaosFEX_{RH25L75G}+$kNN, $ChaosFEX_{RH50L50G}+$kNN and $ChaosFEX_{RH75L25G}+$kNN is seen in Table ~\ref{Table:Macro F1 Score kNN}.

\begin{table}[h!]
\centering
\caption{Macro F1-scores obtained for different ChaosFEX$_{RH}+$kNN architectures.\\{\bf Bold} fonts indicate the highest F1 score achieved for the respective dataset.}
\label{Table:Macro F1 Score kNN}
\begin{tabular}{llll}
\hline
\textbf{Data-set Name} & $ChaosFEX_{RH25L75G}$+kNN & $ChaosFEX_{RH50L50G}$+kNN & $ChaosFEX_{RH75L25G}$+kNN \\ \hline
Iris & \textbf{1} & \textbf{1} &\textbf{1} \\ \hline
Ionosphere & 0.74 & \textbf{0.85} & 0.80 \\ \hline
Wine & 0.66 & 0.72 & \textbf{0.77} \\ \hline
Bank-Note-Authentication & \textbf{0.93} & 0.83 & 0.89 \\ \hline
Haberman's-Survival & \textbf{0.64} & 0.61 & 0.61 \\ \hline
Breast-Cancer-Wisconsin & \textbf{0.98} & 0.93 & 0.94 \\ \hline
Statlog(Heart) & 0.60 & \textbf{0.81} & 0.78 \\ \hline
Seeds & 0.76 & 0.70 & \textbf{0.79} \\ \hline
\end{tabular}%
\end{table}

\begin{table}[h!]
\centering
\caption{Macro F1-scores obtained for different ChaosFEX$_{RH}+$GNB architectures.\\{\bf Bold} fonts indicate the highest F1-score achieved for the respective data-set.}
\label{Table:Macro F1 Score obtained for GNB}
\begin{tabular}{llll}
\hline
\textbf{Data-set Name} & $ChaosFEX_{RH25L75G}$+GNB & $ChaosFEX_{RH50L50G}$+GNB& $ChaosFEX_{RH75L25G}$+GNB \\ \hline
Iris &\textbf{1}  &\textbf{1}  1 & 0.97 \\ \hline
Ionosphere & 0.83 & 0.83 &\textbf{ 0.91} \\ \hline
Wine & \textbf{0.94} & \textbf{0.94} &\textbf{0.94} \\ \hline
Bank-Note-Authentication & \textbf{0.73} & 0.67 & 0.70 \\ \hline
Haberman's-Survival & \textbf{0.62} & 0.61 & 0.52 \\ \hline
Breast-Cancer-Wisconsin & \textbf{0.94} & 0.89 & 0.91 \\ \hline
Statlog(Heart) & 0.77 & \textbf{0.81} & 0.74 \\ \hline
Seeds & \textbf{0.72} & 0.63 & 0.70 \\ \hline
\end{tabular}%
\end{table}
\newpage
%

The macro F1 scores obtained for $ChaosFEX_{RH25L75G}+$GNB, $ChaosFEX_{RH50L50G}+$GNB and $ChaosFEX_{RH75L25G}+$GNB can be found in Table ~\ref{Table:Macro F1 Score obtained for GNB}.




The macro F1 scores obtained for $ChaosFEX_{RH25L75G}+$Random Forests, $ChaosFEX_{RH50L50G}+$Random Forests and $ChaosFEX_{RH75L25G}+$Random Forests can be found in Table ~\ref{Table: Macro F1 RF}. High performance is obtained for {\it Breast Cancer Wisconsin} dataset using $ChaosFEX_{RH25L75G}+$Random Forests with F1 score of $0.98$.

\begin{table}[h!]
\centering
\caption{Macro F1 scores obtained for all datasets using ChaosFEX$_{RH}+$RF architectures.\\{\bf Bold} fonts indicate the highest F1 score achieved for the respective dataset.}
\label{Table: Macro F1 RF}
\begin{tabular}{llll}
\hline
\textbf{Data-set Name} & $ChaosFEX_{RH25L75G}$+RF & $ChaosFEX_{RH50L50G}$+RF & $ChaosFEX_{RH75L25G}$+RF  \\ \hline
Iris & \textbf{1} & \textbf{1} & 0.97 \\ \hline
Ionosphere & 0.96 & 0.93 & \textbf{0.97} \\ \hline
Wine & \textbf{0.97} & \textbf{0.97} & \textbf{0.97} \\ \hline
Bank-Note-Authentication & 0.93 & 0.92 & \textbf{0.94} \\ \hline
Haberman's-Survival & \textbf{0.66} & 0.57 & 0.59 \\ \hline
Breast-Cancer-Wisconsin & 0.98 & \textbf{0.99} & 0.97 \\ \hline
Statlog(Heart) & 0.86 & \textbf{0.87} & 0.71 \\ \hline
Seeds & \textbf{0.83} & 0.76 & 0.78 \\ \hline
\end{tabular}%
\end{table}
 When compared with earlier architectures which were either homogeneous NL~\cite{sethi2023neurochaos} or heterogeneous NL but with fixed structure (odd-even)~\cite{remya2020analysis}, we report that RHNL yields either comparable or superior classification performance. For ease of comparison, we summarize these results in Table~\ref{table:F1 Compare}. Macro F1 scores obtained with RHNL are among the best for the various datasets considered in our study.
\begin{table}[h!]
\caption{Comparison of best macro F1-scores obtained for RHNL structures (proposed in this study) with the other architectures reported in ~\cite{as2023analysis}. Best macro F1 scores are highlighted in {\bf bold} font.}
\label{table:F1 Compare}
\begin{tabular}{p{2cm}p{1cm}p{4.5cm}p{1cm}p{2cm}}
\hline
\textbf{Data-set} & \textbf{Best macro F1 score} & \textbf{RHNL architectures with best macro F1-scores} & \multicolumn{2}{l}{\textbf{NL with best macro F1 scores reported in ~\cite{as2023analysis}}} \\ \hline
\multirow{5}{*}{Iris} & \multirow{5}{*}{\textbf{1}} & $ChaosFEX_{RH25L75G}$, \\&& $ChaosFEX_{RH50L50G}$, \\& & $ChaosFEX_{RH75L25G}$, \\&&$ChaosFEX_{RH25L75G}$+SVM,\\& & $ChaosFEX_{RH50L50G}$+SVM, \\&&$ChaosFEX_{RH25L75G}$+AB, & \multirow{5}{*}{1} & $ChaosFEX_{Logistic}$ \\
 &  & $ChaosFEX_{RH50L50G}$+AB,\\&& $ChaosFEX_{RH525L75G}$+DT,\\&& $ChaosFEX_{RH525L75G}$+kNN, &  & $ChaosFEX_{Logistic}$+SVM \\
 &  & $ChaosFEX_{RH25L75G}$+GNB,\\&& $ChaosFEX_{RH25L75G}$+RF,\\&&$ChaosFEX_{RH50L50G}$+RF, &  & $ChaosFEX_{Hetero}$ \\
 &  & ChaosFEX$_{RH25L75G}+$SVM,\\&& ChaosFEX$_{RH50L50G}+$SVM,\\ &  & ChaosFEX$_{Hetero}+$SVM \\
 &  &ChaosFEX$_{RH75L25G}+$SVM &  &  \\ \hline
\multirow{3}{*}{Ionosphere} & \multirow{3}{*}{\textbf{0.97}} & $ChaosFEX_{RH25L75G}$+AB, \\&&$ChaosFEX_{RH50L50G}$+AB, & \textbf{0.97} & $ChaosFEX_{Logistic}$+SVM \\
 &  & $ChaosFEX_{RH75L25G}$+AB,\\&&$ChaosFEX_{RH75L25G}$+DT, &  &  \\
 &  & $ChaosFEX_{RH75L25G}$+RF &  &  \\ \hline
Wine & \textbf{1} & $ChaosFEX_{RH50L50G}$+AB & 0.98 & $ChaosFEX_{GLS}$ \\ \hline
Bank-Note-Authentication & \textbf{0.99} & $ChaosFEX_{RH50L50G}$+AB,\\&& $ChaosFEX_{RH75L25G}$+AB, & 0.96 & $ChaosFEX_{Logistic}$+SVM \\ \hline
Haberman's-Survival & \textbf{0.73} & $ChaosFEX_{RH25L75G}$ & 0.72 & $ChaosFEX_{Hetero}$ \\ \hline
Breast-Cancer-Wisconsin & \textbf{0.99} & $ChaosFEX_{RH75L25G}$, \\&&$ChaosFEX_{RH50L50G}$+RF & 0.97 & $ChaosFEX_{Logistic}$+SVM\\ \hline
Statlog(Heart) & \textbf{0.92} & $ChaosFEX_{RH25L75G}$+DT, & 0.89 & $ChaosFEX_{Logistic}$+SVM \\ \hline
Seeds & \textbf{0.86} & $ChaosFEX_{RH25L75G}$+AdaBoost,\\&& $ChaosFEX_{RH75L25G}$+SVM & \textbf{0.86} & $ChaosFEX_{Hetero}$ \\ \hline
\end{tabular}%
\end{table}
\\

\subsection{Results Obtained for Time Series Dataset}
\begin{figure}[h!]
    \centering

\begin{tikzpicture}[scale=0.75]
 \begin{axis}[
     title={},
     ybar,
    nodes near coords,
     bar width=0.2cm,
     symbolic x coords={,Cosine,SVM,AB,DT,GNB,kNN,RF,},
     enlarge x limits=0.2,
]
\addplot coordinates{(Cosine,0.88)(SVM,0.97)(AB, 0.41)(DT,0.20)(GNB,0.73)(kNN,0.90)(RF,0.70)};
\end{axis}
\end{tikzpicture}
\caption{Macro F1 scores obtained for {\it FSDD} data set for ChaosFEX$_{RH25L75G}$ with various classifiers.  Classifiers are labeled along the x-axis.}   
\label{figure:RH25L75G_FSDD}
\end{figure}
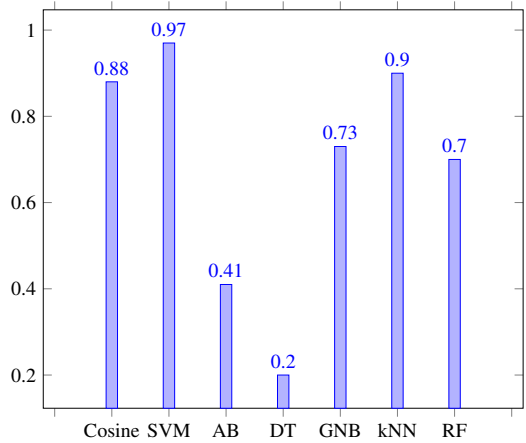

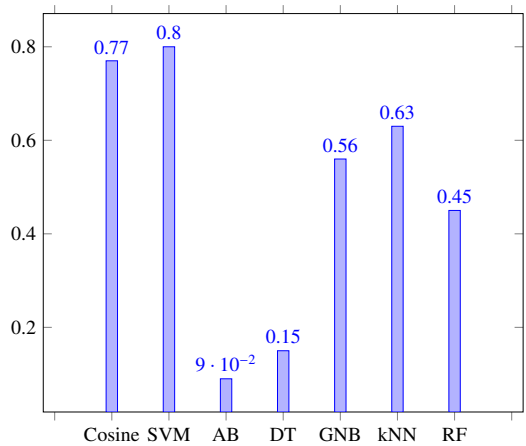
\begin{figure}[h!]
    \centering

\begin{tikzpicture}[scale=0.75]
\begin{axis}[
    title={},
    ybar,
    nodes near coords,
    bar width=0.2cm,
    symbolic x coords={,Cosine,SVM,AB,DT,GNB,kNN,RF,},
    enlarge x limits=0.2,
]
\addplot coordinates{(Cosine,0.77)(SVM,0.80)(AB, 0.09)(DT,0.15)(GNB,0.56)(kNN,0.63)(RF,0.45)};
\end{axis}
\end{tikzpicture}
\caption{Macro F1 scores obtained for {\it FSDD} data set for ChaosFEX$_{RH50L50G}$ with various classifiers.  Classifiers are labeled along the x-axis.} 
\label{figure:RH50L50G_FSDD}
\end{figure}

\begin{figure}[h!]
    \centering
\begin{tikzpicture}[scale=0.75]
\begin{axis}[
    title={},
    ybar,
    nodes near coords,
    bar width=0.2cm,
    symbolic x coords={,Cosine,SVM,AB,DT,GNB,kNN,RF,},
    enlarge x limits=0.2,
]
\addplot coordinates{(Cosine,0.94)(SVM,0.98)(AB, 0.33)(DT,0.61)(GNB,0.61)(kNN,0.84)(RF,0.51)};
\end{axis}
\end{tikzpicture}
\caption{Macro F1 scores obtained for {\it FSDD} data set for ChaosFEX$_{RH75L25G}$ with various classifiers.  Classifiers are labeled along the x-axis.} 
\label{figure:RH75L25G_FSDD}
\end{figure}
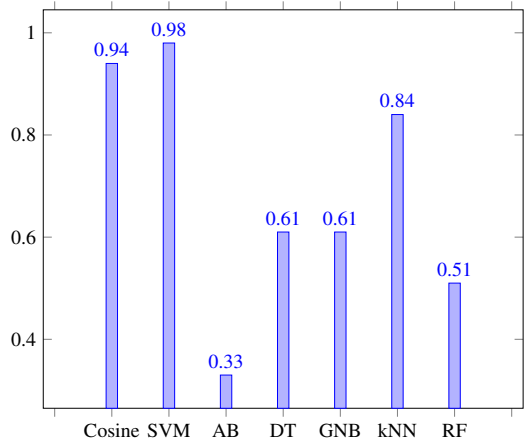

We analysed the performance of our proposed RHNL architectures with a time series dataset -- namely {\it Free Spoken Digit Dataset} (FSDD). The hyperparameters tuned are given in Tables~\ref{table:FSSD 25L75G Hyperparameters},~\ref{table:FSSD 50L50GHyperparameters},~\ref{table:FSSD 75L25G Hyperparameters} and~\ref{table:hyperparameters_FSDD_OtherClassifiers}. Macro F1 scores obtained for various $ChaosFEX_{RHNL}$ architectures are seen in Figures~\ref{figure:RH25L75G_FSDD},~\ref{figure:RH50L50G_FSDD} and~\ref{figure:RH75L25G_FSDD} (respectively).

\subsection{Classification performance of $ChaosFEX_{RHNL}$ for Debris Scars and Urban Images}
\begin{figure}[htbp]
    \centering
    \begin{minipage}[t]{0.45\textwidth}
        \centering
        \includegraphics[width=\linewidth]{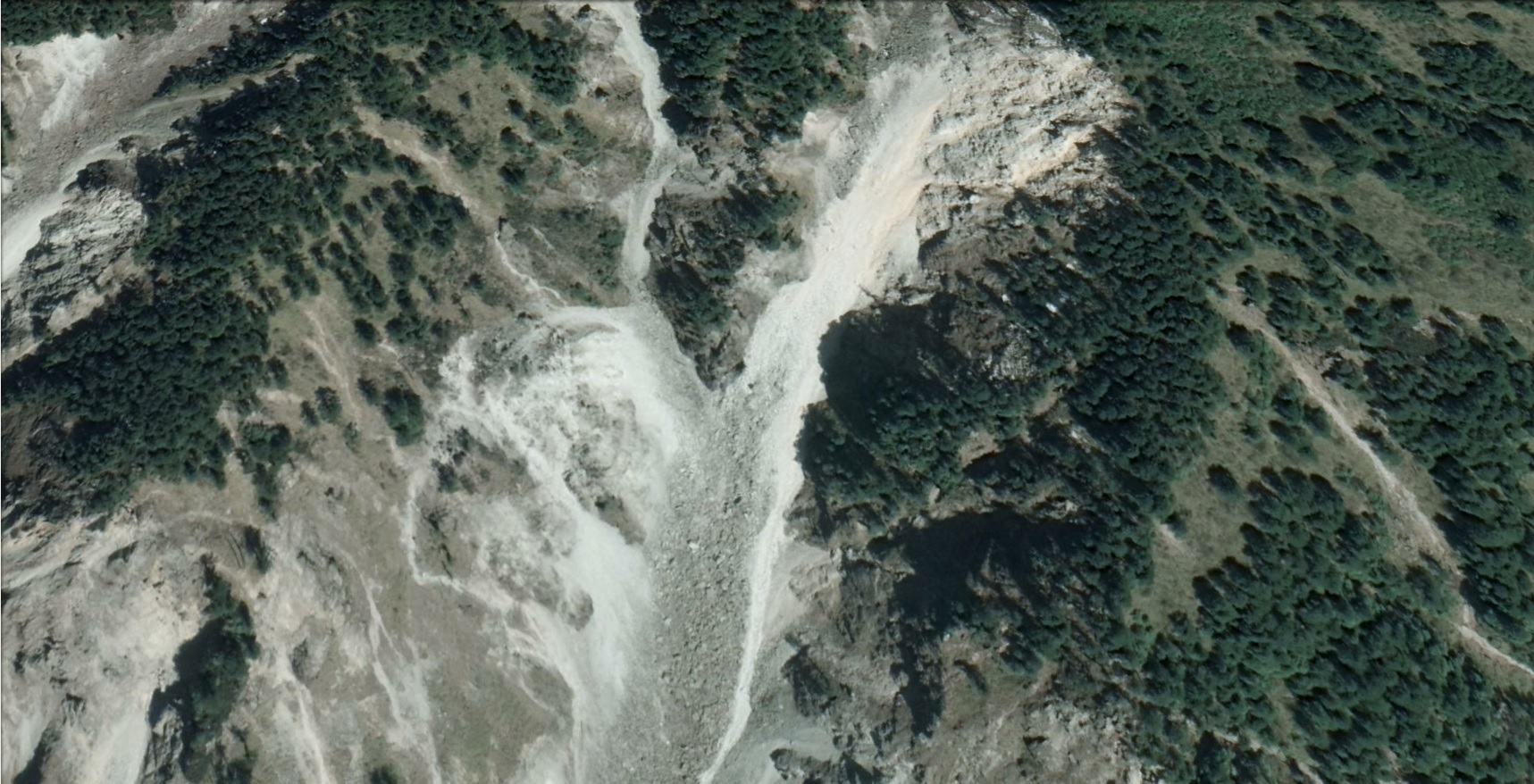}
        (a)
    \end{minipage}
    \hfill
    \begin{minipage}[t]{0.45\textwidth}
        \centering
        \includegraphics[width=\linewidth]{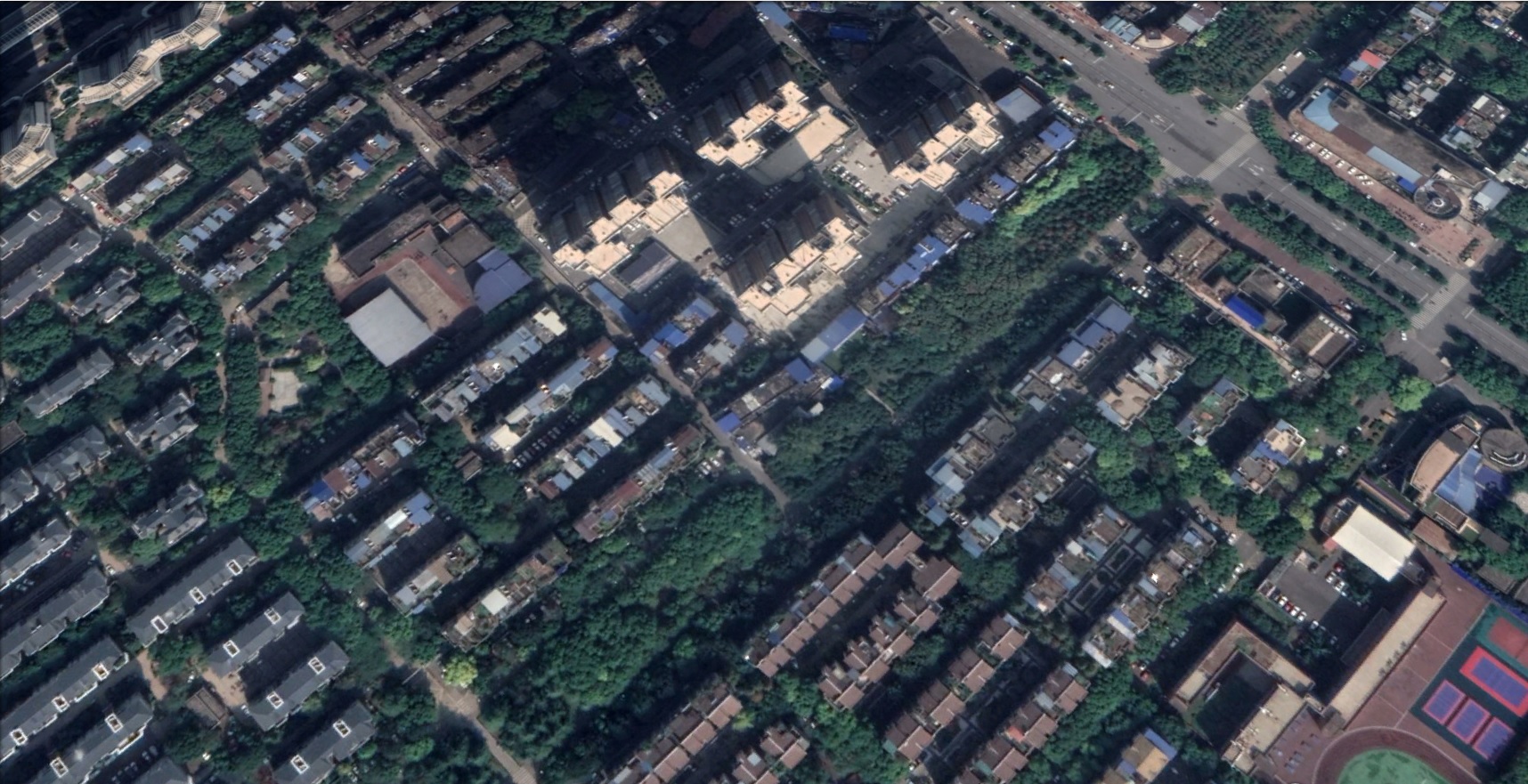}
        (b)
    \end{minipage}
    \caption{(a) Debris scar image (class ``debris''). (b) Urban settlement image (class ``urban'').}
    \label{fig:debrisurbanimages}
\end{figure}

Satellites images are processed to detect and estimate vulnerability of human settlements. Machine Learning algorithms are used now a days to identify areas with high risk of landslide ~\cite{sridharan2020novel}. Around $85$ images of debris scars and urban settlements from five Asian countries (India, Nepal, Japan, Taiwan and China) were obtained from Planet labs~\cite{Planet} imagery with $3-5$m resolution for our analysis to identify the classification performance of Neurochaos Learning Architecture (NL),  specifically ChaosFEX$_{RH25L75G}$ algorithm. Images are labelled either as ``debris'' or as ``urban'' based on visual recognition. We have used $35$ debris scar images and $50$ urban settlement images for our analysis.  Figure~\ref{fig:debrisurbanimages}(a) and Figure~\ref{fig:debrisurbanimages}(b) shows the sample images from class ``debris'' and class ``urban''. Out of total $85$ satellite images captured, $80$\% of the data are used for training and the remaining $20$\% are used for testing. We have used $5-$fold cross validation for our analysis.

The images are initially pre-processed using Otsu global thresholding algorithm and then filtered using Discrete Wavelet Transform(DWT). Daubechies-4 wavelets are used for the DWT implementation.  Grey-level co-occurrence matrix (GLCM) is then created which analyses the pairs of horizontally adjacent pixels in a scaled version of the image. From the GLCM matrix, $12$ features namely Contrast, Correlation, Energy, Homogeneity, Mean, Standard Deviation, Entropy, RMS, Variance, Smoothness, Kurtosis and Skewness are extracted. These features are fed to various $ChaosFEX_{RHNL}$.

Parameters tuned for various structures considered for analysis with debris-urban dataset are shown in Appendix Tables~\ref{table:25L75Gparameters debrisurban},~\ref{table50L50Gparametersdebrisurban} and ~\ref{table75L25Gparametersdebrisurban}.

Performance analysis is done and the macro F1 score obtained for the various architectures considered are given in Table~\ref{F1 DebrisUrban}. We analysed the structures and found that high F1 score of $0.94$ is obtained with $ChaosFEX_{RH25L75G}$, $ChaosFEX_{RH25L75G}$ +SVM, $ChaosFEX_{RH25L75G}$ +kNN, $ChaosFEX_{RH25L75G}$ +DT and $ChaosFEX_{RH25L75G}$+RF for the debris-urban dataset considered.

\begin{table}[h!]
\centering
\caption{Macro F1 scores obtained for debris-urban dataset with various $ChaosFEX_{RHNL}$ combined with various classifiers (cosine similarity and other ML classifiers).}
\label{F1 DebrisUrban}
\resizebox{\textwidth}{!}{%
\begin{tabular}{llllllll}
\hline
\textbf{RHNL with various classifiers} & \textbf{Cosine Similarity} & \textbf{SVM} & \textbf{k-NN} & \textbf{AB} & \textbf{DT} & \textbf{GNB} & \textbf{RF} \\ \hline
$ChaosFEX_{RH25L75G}$ & 0.94 & 0.94 & 0.94 & 0.70 & 0.94 & 0.88 & 0.94 \\ \hline
$ChaosFEX_{RH50L50G}$ & 0.70 & 0.70 & 0.74 & 0.81 & 0.82 & 0.80 & 0.88 \\ \hline
$ChaosFEX_{RH75L25G}$ & 0.66 & 0.94 & 0.70 & 0.88 & 0.88 & 0.87 & 0.88 \\ \hline
\end{tabular}%
}
\end{table}

\subsection{ Classification performance of $ChaosFEX_{RHNL}$ for Brain Tumor Dataset}
Brain tumors are the leading cause of cancer death in children. They are caused by the abnormal and uncontrolled growth of cells inside the brain or spinal canal. Classification of brain tumors using machine learning technology is very relevant for radiologists to confirm their analysis more effectively and quickly. 

\begin{figure}[htbp]
    \centering
    \begin{minipage}[t]{0.45\textwidth}
        \centering
        \includegraphics[scale=0.68]{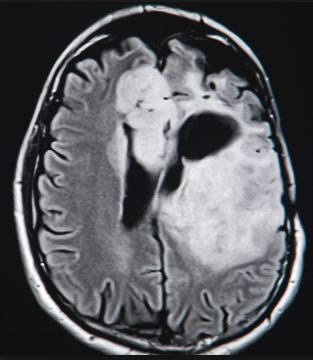}
        (a)
    \end{minipage}
    \hfill
    \begin{minipage}[t]{0.45\textwidth}
        \centering
        \includegraphics[width=\linewidth]{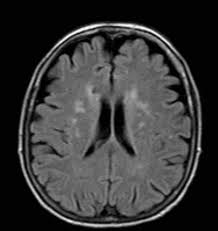}
        (b)
    \end{minipage}
    \caption{(a) MRI image showing malignant brain tumour (class ``malignant''). (b) MRI image showing benign brain tumour (class ``benign''). }
    \label{MRIimage_examples}
\end{figure}

For our analysis, we have considered $100$ MRI brain images from {\it Kaggle} online repository~\cite{Kaggle}. Images are labelled as ``malignant'' or ``benign''. We considered $40$ malignant and benign images for our analysis. We split $80$\% of data for training and the remaining $20$\% for testing. Five-fold cross validation is adopted in this analysis. Figure~\ref{MRIimage_examples}(a) and Figure~\ref{MRIimage_examples}(b) shows sample images from each of the two classes (malignant and benign).
 
Preprocessing is done by applying anisotropic filtering of all images. Grey-level co-occurrence matrix (GLCM) is then created which analyses the pairs of horizontally adjacent pixels in a scaled version of the image. From the GLCM matrix, we can calculate the features needed for classification. The twelve features generated are {\it Contrast, Correlation, Energy, Homogeneity, Mean, Standard Deviation, Entropy, RMS, Variance, Smoothness, Kurtosis} and {\it Skewness}.  These extracted features are subsequently fed to stand-alone SVM, stand-alone k-NN, and ChaosFEX$_{RH25L75G}$ +SVM and ChaosFEX$_{RH25L75G}$ +k-NN. The hyperparameters tuned for various architectures under analysis for brain tumor dataset is given in Appendix Tables~\ref{table:parameter25L75GBrainTumor},~\ref{table:hyperparameters50L50Gbraintumor} and ~\ref{table:parameters75L25GBT}.

\begin{table}[]
\caption{
Macro F1 scores obtained for MRI brain tumour dataset with various $ChaosFEX_{RHNL}$ combined with various classifiers (cosine similarity and other ML classifiers).}
\label{table:F1ScoreBrainTumor}
\resizebox{\textwidth}{!}{%
\begin{tabular}{llllllll}
\hline
\textbf{RHNL with various classifiers} & \textbf{Cosine Similarity} & \textbf{SVM} & \textbf{k-NN} & \textbf{AB} & \textbf{DT} & \textbf{GNB} & \textbf{RF} \\ \hline
$ChaosFEX_{RH25L75G}$& 0.881 & 0.812 & 0.72 & 0.76 & 0.81 & 0.48 & 0.83 \\ \hline
$ChaosFEX_{RH50L50G}$ & 0.44 & 0.50 & 0.69 & 0.78 & 0.73 & 0.78 & 0.73 \\ \hline
$ChaosFEX_{RH75L25G}$ & 0.83 & 0.78 & 0.76 & 0.82 & 0.60 & 0.50 & 0.73 \\ \hline
\end{tabular}%
}
\end{table}

The results (table~\ref{table:F1ScoreBrainTumor}) show classification performance is better with $ChaosFEX_{RH25L75G}$ (F1 score $=0.881$) for brain tumor dataset. We may further improve the classification performance with other ChaosFEX$_{RHNL}$ architectures with properly tuned hyperparameters.

\subsection{Performance of $ChaosFEX_{RHNL}$ compared with Stand-alone ML Classifiers}
When compared with either homogeneous NL or heterogeneous NL but with fixed structure (odd-even placement of GLS and Logistic neurons), we have reported that RHNL yields either comparable or superior classification performance. 
Table~\ref{table:RHNL SA Compare} compares the highest F1 score obtained with various stand-alone ML classifiers (SA-ML) and RHNL architectures for all the $11$ datasets in this study. For \textit{Iris, Wine} and \textit{Bank Note Authentication} datasets, RHNL architectures perform equally well with some of the stand-alone ML classifiers. It is interesting to note that a $16$\% increase in performance is achieved for {\it Haberman's Survival} dataset with our proposed RHNL structure when compared with GNB which gives the best F1 score among all the stand-alone ML classifiers (F1-score $=0.57$). Also for \textit{Ionosphere, Breast Cancer Wisconsin} and \textit{Statlog (Heart)} datasets, our newly proposed RHNL structures outperform stand-alone classifiers significantly well. However for \textit{Seeds} dataset alone, RHNL gives a lower performance. Table~\ref{table:sa comparison images} shows that $ChaosFEX_{RH25L75G}$, $ChaosFEX_{RH25L75G}$+SVM, $ChaosFEX_{RH25L75G}$+kNN, $ChaosFEX_{RH25L75G}$+DT and $ChaosFEX_{RH25L75G}$+RF give higher performance (F1 Score $= 0.94$) than any of the stand-alone ML classifiers for the debris-urban dataset. For brain tumor dataset also $ChaosFEX_{RH25L75G}$ outperforms all the stand-alone ML classifiers.

\begin{table}[h!]
\caption{Comparison of RHNL architecture with the best stand-alone ML classifiers. Only the standalone ML classifier which yielded the highest F1-score is mentioned. (SA: Standalone). }
\label{table:RHNL SA Compare}
\resizebox{\textwidth}{!}{%
\begin{tabular}{lccc}
\hline
Dataset & Highest F1-score for SA-ML & SA-ML & Highest F1-score with RHNL \\ \hline
Iris & 1.00 & RF, k-NN & 1.00 \\ \hline
Ionosphere & 0.96 & SVM & 0.97 \\ \hline
Wine & 1.00 & GNB & 1.00 \\ \hline
Bank Note Authentication & 0.99 & SVM, k-NN & 0.99 \\ \hline
Haberman’s Survival & 0.57 & GNB & 0.73 \\ \hline
Breast Cancer Wisconsin & 0.95 & k-NN & 0.99 \\ \hline
Statlog(Heart) & 0.84 & k-NN, SVM & 0.92 \\ \hline
Seeds & 0.92 & k-NN & 0.86 \\ \hline
FSDD & 0.97 & RF & 0.98 \\ \hline
\end{tabular}%
}
\end{table}

\begin{table}[h!]
\caption{Comparison of the performance (F1-scores) of $ChaosFEX_{RH25L75G}$ architecture with stand-alone ML classifiers for the {\it debris-urban} and {\it brain tumor image} datasets. SA: Stand alone ML classifiers. RHNL gives the best F1-scores (emphasized in bold).}

\label{table:sa comparison images}
\resizebox{\textwidth}{!}{%
\begin{tabular}{llllllll}
\hline
Dataset & SA-SVM & SA-kNN & SA-AB & SA-DT & SA-GNB & SA-RF & $ChaosFEX_{RH25L75G}$ \\ \hline
Debris-Urban & 0.71 & 0.71 & 0.71 & 0.71 & 0.71 & 0.82 & \textbf{0.94} \\ \hline
Brain Tumor & 0.7 & 0.6 & 0.70 & 0.7 & 0.55 & 0.75 & \textbf{0.88} \\ \hline
\end{tabular}%
}
\end{table}
\begin{table}[h!]
\caption{Performance comparison of $ChaosFEX_{RHNL}$+ML structures proposed in this study with stand-alone ML classifiers. Checkmark~(\checkmark) indicates that the algorithm gives highest F1 score (value reported in the second column). As it can be seen RHNL yields the best performance in $10$ out of $11$ datasets.}
\label{table:StandaloneRHNL Comparison}
\resizebox{\textwidth}{!}{%
\begin{tabular}{cccc}
\hline
\textbf{Dataset} & \textbf{Highest         Macro F1 Score} & \textbf{Stand-alone ML classifiers} & \textbf{$ChaosFEX_{RHNL}$+ML Structures} \\ \hline
Iris & 1.00 & \checkmark & \checkmark \\ \hline
Ionosphere & 0.96 &  & \checkmark \\ \hline
Wine & 1.00 &\checkmark & \checkmark \\ \hline
Bank Note Authentication & 0.99 & \checkmark & \checkmark \\ \hline
Haberman's Survival & 0.73 &  & \checkmark \\ \hline
Breast Cancer Wisconsin & 0.99 & & \checkmark \\ \hline
Statlog Heart & 0.92 & & \checkmark \\ \hline
Seed & 0.86 & \checkmark &  \\ \hline
FSDD & 0.98 & &  \checkmark \\ \hline
Debris-Urban Image & 0.94 & & \checkmark \\ \hline
Brain Tumor Image & 0.83 &  & \checkmark \\ \hline
\end{tabular}%
}
\end{table}
Table~\ref{table:StandaloneRHNL Comparison} shows that highest macro F1 score is obtained for $ChaosFEX_{RHNL}$ structures for all dataset except \textit{Seed}.

\subsection{Performance of RHNL in low training sample regime}

One of the major significance of Neurochaos Learning architectures is that they perform well in the low training sample regime~\cite{balakrishnan2019chaosnet}. We analysed the performance of ChaosFEX$_{RH25L75G}$ for the {\it MRI brain tumor} dataset in the low training sample regime and compared its peformance with the standalone ML classifiers. 
\begin{figure}[h!]
			\centering
    		\includegraphics[scale=0.4]{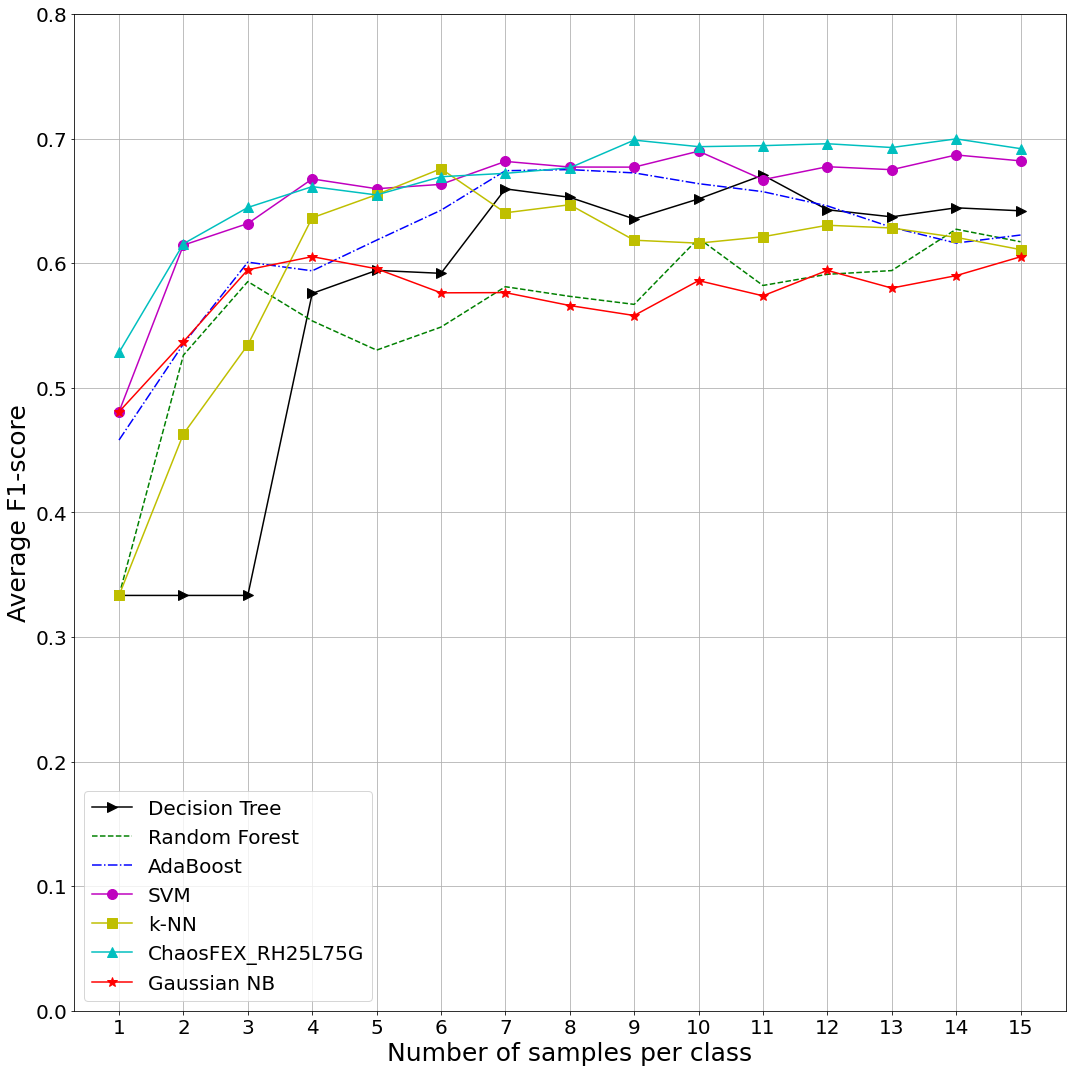}
  			\caption{\label{figure1_1:LowSampleRegimeRHNL_Performance}Comparative performance
of ChaosFEX$_{RHNL25L75G}$ with stand-alone ML classifiers in the regime of low number of training instances. }
  		\end{figure}
The $12$ features generated from the selected $100$ brain tumor MRI images from {\it Kaggle} dataset online repository~\cite{Kaggle} are used for our analysis. Analysis is done starting with one sample per class in the training set and the remaining samples were used for testing. The process is repeated with $2$ to $15$ samples per class in train data set and the remaining samples in test dataset. In every case, we did $10$ indepedent random trials. The average F1 score of these $10$ trials are reported in each case and compared with the results obtained for stand-alone classifiers namely Decision Tree (DT), Random Forest (RF), AdaBoost (AB), SVM, k-NN, Guassian Naive Bayes (GNB) and ChaosFEX$_{RH25L75G}$. In ChaosFEX$_{RH25L75G}$, $25$\% of the locations are randomly allotted to logistic map neurons and the remaining locations with GLS neurons. Cosine Similarity classifier is used in ChaosFEX$_{RH25L75G}$ architecture. Figure~\ref{figure1_1:LowSampleRegimeRHNL_Performance} shows that for low training samples, ChaosFEX$_{RH25L75G}$ consistently high performance with respect to other standalone ML algorithms.

For training dataset with one sample per class, ChaosFEX$_{RHNL25L75G}$ gives high performance and Decision Tree and K-NN gives low performance. For training dataset with $4$ and $5$ samples per class, SVM outperforms ChaosFEX$_{RH25L75G}$. k-NN gives high performance with $6$ samples per class in training dataset. However from $8$ samples per class in train dataset onwards, ChaosFEX$_{RHNL25L75G}$ performed well again and continues to outperform other classifiers. This shows that RHNL architecture is able to learn with very few training samples per class making it very desirable in practical applications where there is a paucity of training data.

\newpage
\section{Conclusion and Future Research Directions}
\label{Conclusions}
Incorporation of both {\it heterogeneity} and {\it randomness} into Neurochaos Learning is one step in the right direction of mimicking the complex neural organization of the human brain. In this study, we find that such Random Heterogeneous Neurochaos Learning (RHNL) architectures perform very well. When compared with earlier architectures which were either homogeneous NL~\cite{sethi2023neurochaos} or heterogeneous NL but with fixed structure (odd-even)~\cite{remya2020analysis}, we report that RHNL yields either comparable or superior classification performance.Table~\ref{table:F1 Compare} shows that the macro F1 scores obtained with RHNL are among the best for all the datasets considered in our study.
 RHNL architectures beat both homogeneous NL and fixed heterogeneous NL architectures on nearly all classification tasks. RHNL achieved a high level of F1 score for {\it Wine} dataset ($1.0$), {\it Bank Note Authentication} dataset ($0.99$), {\it Breast Cancer Wisconsin} dataset ($0.99$) and {\it FSDD} dataset ($0.98$). RHNL is also superior on time series dataset. For the {\it FSDD} dataset, $ChaosFEX_{RH75L25G}$+SVM achieves the highest macro F1 score of $0.98$. This is infact higher than the best F1 score obtained with homogeneous NL architecture reported in ~\cite{sethi2023neurochaos}. These results of RHNL are clearly among  the best in the literature on the same datasets.
For the debris-urban image data, $ChaosFEX_{RH25L75G}$ gave a high F1 score of $0.94$ with cosine similarity, SVM, k-NN, Decisrion Tree and Randim Forest classifiers. Brain tumor image dataset gives high F1 score of $0.881$ with $ChaosFEX_{RH25L75G}$. RHNL also yields best F1 score when compared with standalone-ML classifiers except for the {\it Seeds} dataset. In the case of low training sample regime, RHNL outperforms almost all traditional ML classifiers.

This line of work can be applied to classification of other datasets in various application domains. For future research, we will explore incorporating other chaotic maps such as  Standard map, Circle map, Ikeda map, H\'enon map, Gumowski-Mira map, Arnold's cat map,  Lorenz system, Baker's map, Lozi map, Hindmarsh-Rose neuronal model, R\"ossler system and other such dynamical systems as neurons in RHNL.

\section{Acknowledgements}
Authors are thankful for the help received by Harikrishnan NB and Deeksha S with Python programs of homogeneous Neurochaos Learning architecture. RA is grateful to the computing facilities provided by Amrita Vishwa Vidyapeetham (Amritapuri Campus). 
\bibliographystyle{plain} 

\begin{thebibliography}{10}
\expandafter\ifx\csname url\endcsname\relax
  \def\url#1{\texttt{#1}}\fi
\expandafter\ifx\csname urlprefix\endcsname\relax\def\urlprefix{URL }\fi
\expandafter\ifx\csname href\endcsname\relax
  \def\href#1#2{#2} \def\path#1{#1}\fi

\bibitem{ramachandran1998phantoms}
V.~Ramachandran, S.~Blakeslee, R.~J. Dolan, Phantoms in the brain probing the mysteries of the human mind, Nature 396~(6712) (1998) 639--640.

\bibitem{graves2013speech}
A.~Graves, A.-r. Mohamed, G.~Hinton, Speech recognition with deep recurrent neural networks, in: 2013 IEEE international conference on acoustics, speech and signal processing, Ieee, 2013, pp. 6645--6649.

\bibitem{harikrishnan2018machine}
N.~Harikrishnan, R.~Vinayakumar, K.~Soman, A machine learning approach towards phishing email detection, in: Proceedings of the Anti-Phishing Pilot at ACM International Workshop on Security and Privacy Analytics (IWSPA AP), Vol. 2013, 2018, pp. 455--468.

\bibitem{sebe2005machine}
N.~Sebe, Machine learning in computer vision, Vol.~29, Springer Science \& Business Media, 2005.

\bibitem{harikrishnan2019vision}
J.~Harikrishnan, A.~Sudarsan, A.~Sadashiv, R.~A. Ajai, Vision-face recognition attendance monitoring system for surveillance using deep learning technology and computer vision, in: 2019 international conference on vision towards emerging trends in communication and networking (ViTECoN), IEEE, 2019, pp. 1--5.

\bibitem{remya2020analysis}
A.~Remya~Ajai, S.~Gopalan, Analysis of active contours without edge-based segmentation technique for brain tumor classification using svm and knn classifiers, in: Advances in Communication Systems and Networks: Select Proceedings of ComNet 2019, Springer, 2020, pp. 1--10.

\bibitem{krishna2019analysis}
S.~Krishna, A.~R. Ajai, Analysis of three point checklist and abcd methods for the feature extraction of dermoscopic images to detect melanoma, in: 2019 9th International Symposium on Embedded Computing and System Design (ISED), IEEE, 2019, pp. 1--5.

\bibitem{asif2023enhanced}
S.~Asif, M.~Zhao, F.~Tang, Y.~Zhu, An enhanced deep learning method for multi-class brain tumor classification using deep transfer learning, Multimedia Tools and Applications (2023) 1--28.

\bibitem{aihara1990chaotic}
K.~Aihara, T.~Takabe, M.~Toyoda, Chaotic neural networks, Physics letters A 144~(6-7) (1990) 333--340.

\bibitem{delahunt2019putting}
C.~B. Delahunt, J.~N. Kutz, Putting a bug in ml: The moth olfactory network learns to read mnist, Neural Networks 118 (2019) 54--64.

\bibitem{balakrishnan2019chaosnet}
H.~N. Balakrishnan, A.~Kathpalia, S.~Saha, N.~Nagaraj, Chaosnet: A chaos based artificial neural network architecture for classification, Chaos: An Interdisciplinary Journal of Nonlinear Science 29~(11) (2019) 113125.

\bibitem{nb2022causality}
H.~NB, A.~Kathpalia, N.~Nagaraj, Causality preserving chaotic transformation and classification using neurochaos learning, Advances in Neural Information Processing Systems 35 (2022) 2046--2058.

\bibitem{korn2003there}
H.~Korn, P.~Faure, Is there chaos in the brain? ii. experimental evidence and related models, Comptes rendus biologies 326~(9) (2003) 787--840.

\bibitem{perez2021neural}
N.~Perez-Nieves, V.~C. Leung, P.~L. Dragotti, D.~F. Goodman, Neural heterogeneity promotes robust learning, Nature communications 12~(1) (2021) 5791.

\bibitem{Weis2019}
S.~Weis, M.~Sonnberger, A.~Dunzinger, E.~Voglmayr, M.~Aichholzer, R.~Kleiser, P.~Strasser, \href{https://doi.org/10.1007/978-3-7091-1544-2_10}{Histological Constituents of the Nervous System}, Springer Vienna, Vienna, 2019, pp. 225--265.
\newblock \href {https://doi.org/10.1007/978-3-7091-1544-2_10} {\path{doi:10.1007/978-3-7091-1544-2_10}}.
\newline\urlprefix\url{https://doi.org/10.1007/978-3-7091-1544-2_10}

\bibitem{harikrishnan2020neurochaos}
N.~Harikrishnan, N.~Nagaraj, Neurochaos inspired hybrid machine learning architecture for classification, in: 2020 International Conference on Signal Processing and Communications (SPCOM), IEEE, 2020, pp. 1--5.

\bibitem{sethi2023neurochaos}
D.~Sethi, N.~Nagaraj, N.~Harikrishnan, Neurochaos feature transformation for machine learning, Integration (2023).

\bibitem{harikrishnan2021noise}
N.~Harikrishnan, N.~Nagaraj, When noise meets chaos: Stochastic resonance in neurochaos learning, Neural Networks 143 (2021) 425--435.

\bibitem{as2023analysis}
R.~A. AS, N.~Harikrishnan, N.~Nagaraj, Analysis of logistic map based neurons in neurochaos learning architectures for data classification, Chaos, Solitons \& Fractals 170 (2023) 113347.

\bibitem{nagaraj4unreasonable}
N.~Nagaraj, The unreasonable effectiveness of the chaotic tent map in engineering applications, Chaos Theory and Applications 4~(4)  197--204.

\bibitem{phatak1995logistic}
S.~Phatak, S.~S. Rao, Logistic map: A possible random-number generator, Physical review E 51~(4) (1995) 3670.

\bibitem{alligood1998chaos}
K.~T. Alligood, T.~D. Sauer, J.~A. Yorke, D.~Chillingworth, Chaos: an introduction to dynamical systems, SIAM Review 40~(3) (1998) 732--732.

\bibitem{fisher1936use}
R.~A. Fisher, The use of multiple measurements in taxonomic problems, Annals of eugenics 7~(2) (1936) 179--188.

\bibitem{dua2017uci}
D.~Dua, C.~Graff, et~al., Uci machine learning repository (2017).

\bibitem{sigillito1989classification}
V.~G. Sigillito, S.~P. Wing, L.~V. Hutton, K.~B. Baker, Classification of radar returns from the ionosphere using neural networks, Johns Hopkins APL Technical Digest 10~(3) (1989) 262--266.

\bibitem{vandeginste1990parvus}
B.~Vandeginste, Parvus: An extendable package of programs for data exploration, classification and correlation, m. forina, r. leardi, c. armanino and s. lanteri, elsevier, amsterdam, 1988, price: Us \$\$\$645 isbn 0-444-43012-1, Journal of Chemometrics 4~(2) (1990) 191--193.

\bibitem{gillich2010banknote}
E.~Gillich, V.~Lohweg, Banknote authentication, 1. Jahreskolloquium Bild. Der Autom (2010) 1--8.

\bibitem{haberman1973analysis}
S.~J. Haberman, The analysis of residuals in cross-classified tables, Biometrics (1973) 205--220.

\bibitem{street1993nuclear}
W.~N. Street, W.~H. Wolberg, O.~L. Mangasarian, Nuclear feature extraction for breast tumor diagnosis, in: Biomedical image processing and biomedical visualization, Vol. 1905, SPIE, 1993, pp. 861--870.

\bibitem{jackson2018jakobovski}
Z.~Jackson, C.~Souza, J.~Flaks, Y.~Pan, H.~Nicolas, A.~Thite, Jakobovski/free-spoken-digit-dataset: v1. 0.8, Zenodo, August (2018).

\bibitem{Planet}
Planet, \href{www.planet.com}{{Planet labs}}.
\newline\urlprefix\url{www.planet.com}

\bibitem{Kaggle}
N.~Chakrabarty, \href{https://www.kaggle.com/datasets/navoneel/brain-mri-images-for-brain-tumor-detection/code}{Brain mri images for brain tumor detection}.
\newline\urlprefix\url{https://www.kaggle.com/datasets/navoneel/brain-mri-images-for-brain-tumor-detection/code}

\bibitem{boser1992training}
B.~E. Boser, I.~M. Guyon, V.~N. Vapnik, A training algorithm for optimal margin classifiers, in: Proceedings of the fifth annual workshop on Computational learning theory, 1992, pp. 144--152.

\bibitem{schapire2013explaining}
R.~E. Schapire, Explaining adaboost, in: Empirical Inference: Festschrift in Honor of Vladimir N. Vapnik, Springer, 2013, pp. 37--52.

\bibitem{quinlan1986induction}
J.~R. Quinlan, Induction of decision trees, Machine learning 1 (1986) 81--106.

\bibitem{berrar2018bayes}
D.~Berrar, Bayes’ theorem and naive bayes classifier, Encyclopedia of bioinformatics and computational biology: ABC of bioinformatics 403 (2018) 412.

\bibitem{cover1967nearest}
T.~Cover, P.~Hart, Nearest neighbor pattern classification, IEEE transactions on information theory 13~(1) (1967) 21--27.

\bibitem{breiman2001random}
L.~Breiman, Random forests, Machine learning 45 (2001) 5--32.

\bibitem{sridharan2020novel}
A.~Sridharan, R.~A. AS, S.~Gopalan, A novel methodology for the classification of debris scars using discrete wavelet transform and support vector machine, Procedia computer science 171 (2020) 609--616.

\end{thebibliography}
 
 \newpage
\newpage
\section*{Appendix}
This section contains the additional supplementary details related to the main manuscript. It contains the hyperparameter tuned values for each dataset used in this study, for different classifiers namely AdaBoost, Decision Trees, kNN and Random Forests that were used in  $ChaosFEX_{RH}$ architectures. The hyperparameters tuned for {\it FSSD} dataset for various $ChaosFEX_{RH}$ architectures are also included in this section.
%
%
%
\begin{table}[h!]
\centering
\caption{Tuned hyperparameters for ChaosFEX$_{RH25L75G}+$AdaBoost.
\label{table:25L75G:AdaBoost Hyperparameters}}
\begin{tabular}{lccccc}
\hline

Data-set & \begin{tabular}[c]{@{}l@{}}$q$\end{tabular} & \begin{tabular}[c]{@{}l@{}}$b$\end{tabular} & \begin{tabular}[c]{@{}l@{}}$\epsilon$\end{tabular} & $n\_estimators$

 \\ \hline

Iris & .062 & .185 & .298 & 10  \\ \hline
Ionosphere & .01 & .409 & .051 & 50  \\ \hline
Wine & .46 & .469 & .141 & 10 \\ \hline
Bank-Note-Authentication & .36 & .419 & .121 & 10  \\ \hline
Haberman's-Survival & .05 & .269 & .031 & 5000  \\ \hline
Breast-Cancer-Wisconsin & .17 & .46 & .05 & 100  \\ \hline
Statlog(Heart) & .47 & .489 & .0309 & 5000 \\ \hline
Seeds & .05 & .189 & 0.161 & 50  \\ \hline
\end{tabular}%
\end{table}

\begin{table}[h!]
\centering
\caption{Tuned hyperparameters for ChaosFEX$_{RH50L50G}+$AdaBoost.
\label{table:50L50G AdaBoost Hyperparameters}}
\centering
\begin{tabular}{llllll}
\hline
Data-set & \begin{tabular}[c]{@{}l@{}}$q$\end{tabular} & \begin{tabular}[c]{@{}l@{}}$b$\end{tabular} & \begin{tabular}[c]{@{}l@{}}$\epsilon$\end{tabular} & $n\_estimators$ \\ \hline
Iris & .05 & .359 & .221 & 10 \\ \hline
Ionosphere & .099 & .479 & .061 & 1000 \\ \hline
Wine & .46 & .469 & .131 & 10  \\ \hline
Bank-Note-Authentication & .09 & .289 & .041 & 1000 \\ \hline
Haberman's-Survival & .14 & .489 & .021 & 1000 \\ \hline
Breast-Cancer-Wisconsin & .069 & .139 & .041 & 10  \\ \hline
Statlog(Heart) & .18 & .169 & .011 & 100 \\ \hline
Seeds & .05 & .139 & .151 & 10  \\ \hline
\end{tabular}%
\end{table}

\begin{table}[h!]
\centering
\caption{Tuned hyperparameters for ChaosFEX$_{RH75L25G}+$AdaBoost.}
\label{table:75L25G:AdaBoost}
\centering
\begin{tabular}{llllll}
\hline
Data-set & \begin{tabular}[c]{@{}l@{}}$q$\end{tabular} & \begin{tabular}[c]{@{}l@{}}$b$\end{tabular} & \begin{tabular}[c]{@{}l@{}}$\epsilon$\end{tabular} & $n\_estimators$ \\ \hline
Iris & .15 & .299 & .231 & 10  \\ \hline
Ionosphere & .02 & .219 & .809 & 50 \\ \hline
Wine & .47 & .479 & .131 & 10 \\ \hline
Bank-Note-Authentication & .01 & .259 & .071 & 5000 \\ \hline
Haberman's-Survival & .23 & .1 & .011 & 5000 \\ \hline
Breast-Cancer-Wisconsin & .14 & .489 & .021 & 1000  \\ \hline
Statlog(Heart) & .13 & .1 & .051 & 10 \\ \hline
Seeds & .05 & .189 & .151 & 50 \\ \hline
\end{tabular}%
\end{table}

\begin{table}[h!]
\centering`
\caption{Tuned hyperparameters for ChaosFEX$_{RH25L75G}+$ Decision Trees. 
\label{table:25L75G:Decision Tree}}
\resizebox{\textwidth}{!}{%
\begin{tabular}{llllllll}
\hline
Dataset & \begin{tabular}[c]{@{}l@{}}$q$\end{tabular} & \begin{tabular}[c]{@{}l@{}}$b$\end{tabular} & \begin{tabular}[c]{@{}l@{}}$\epsilon$\end{tabular} & $min\_samples\_leaf$ & $max\_depth$ & $ccp\_alpha$  \\ \hline
Iris & .062 & .185 & .298 & 1 & 2 & 0.0  \\ \hline
Ionosphere & .01 & .409 & .051 & 3 & 7 & 0.0  \\ \hline
Wine & .46 & .469 & .141 & 1 & 4 & 0.0  \\ \hline
Bank-Note-Authentication & .36 & .419 & .121 & 1 & 6 & 0.0  \\ \hline
Haberman's-Survival & .05 & .269 & .031 & 7 & 4 & 0.0  \\ \hline
Breast-Cancer-Wisconsin & .17 & .46 & .05 & 10 & 4 & .0022  \\ \hline
Statlog(Heart) & .47 & .489 & .0309 & 1 & 3 & 0.0  \\ \hline
Seeds & .05 & .189 & .161 & 2 & 3 & 0.0  \\ \hline
\end{tabular}%
}
\end{table}
\begin{table}[h!]
\centering
\caption{Tuned hyperparameters for ChaosFEX$_{RH50L50G}+$ Decision Trees.}
\label{table:50L50G:Decision Tree}
\resizebox{\textwidth}{!}{%
\begin{tabular}{llllllll}
\hline
Data-set & \begin{tabular}[c]{@{}l@{}}$q$\end{tabular} & \begin{tabular}[c]{@{}l@{}}$b$\end{tabular} & \begin{tabular}[c]{@{}l@{}}$\epsilon$\end{tabular} & $min\_samples\_leaf$ & $max\_depth$ & $ccp\_alpha$  \\ \hline
Iris & .05 & .359 & .221 & 1 & 2 & 0.0  \\ \hline
Ionosphere & .099 & .479 & .061 & 3 & 4 & 0.0  \\ \hline
Wine & .46 & .469 & .131 & 1 & 4 & 0.0  \\ \hline
Bank-Note-Authentication & .09 & .289 & .041 & 1 & 8 & 0.0  \\ \hline
Haberman's-Survival & .14 & .489 & .021 & 4 & 5 & 0.0  \\ \hline
Breast-Cancer-Wisconsin & .069 & .139 & .041 & 3 & 3 & 0.0  \\ \hline
Statlog(Heart) & .18 & .169 & .011 & 1 & 4 & 0.0 \\ \hline
Seeds & .05 & .139 & .151 & 1 & 3 & 0.0  \\ \hline
\end{tabular}%
}
\end{table}

\begin{table}[h!]
\centering
\caption{Tuned hyperparameters for ChaosFEX$_{RH75L25G}+$ Decision Trees.}
\label{table:75L25G DecisionTree}
\resizebox{\textwidth}{!}{%
\begin{tabular}{llllllll}
\hline
Data-set & \begin{tabular}[c]{@{}l@{}}$q$\end{tabular} & \begin{tabular}[c]{@{}l@{}}$b$\end{tabular} & \begin{tabular}[c]{@{}l@{}}$\epsilon$\end{tabular} & $min\_samples\_leaf$ & $max\_depth$ & $ccp\_alpha$  \\ \hline
Iris & .15 & .299 & .231 & 1 & 2 & 0.0  \\ \hline
Ionosphere & .02 & .219 & .809 & 1 & 6 & 0.0  \\ \hline
Wine & .47 & .479 & .131 & 2 & 4 & 0.0  \\ \hline
Bank-Note-Authentication & .01 & .259 & .071 & 8 & 7 & 0.0  \\ \hline
Haberman's-Survival & .23 & .1 & .011 & 5 & 6 & 0.0  \\ \hline
Breast-Cancer-Wisconsin & .14 & .489 & .021 & 6 & 9 & .00218  \\ \hline
Statlog(Heart) & .13 & .1 & .051 & 1 & 3 & 0.004  \\ \hline
Seeds & .05 & .189 & .151 & 1 & 4 & 0.0  \\ \hline
\end{tabular}%
}
\end{table}

\begin{table}[h!]
\centering
\caption{Tuned hyperparameters for ChaosFEX$_{RH25L75G}+$kNN.} 
\label{table:25L75G:knn }
\centering
\begin{tabular}{llllll}
\hline
Data-set & \begin{tabular}[c]{@{}l@{}}$q$\end{tabular} & \begin{tabular}[c]{@{}l@{}}$b$\end{tabular} & \begin{tabular}[c]{@{}l@{}}$\epsilon$\end{tabular} & $k$  \\ \hline
Iris & .062 & .185 & .298 & 3 \\ \hline
Ionosphere & .01 & .409 & .051 & 3 \\ \hline
Wine & .46 & .469 & .141 & 5 \\ \hline
Bank-Note-Authentication & .36 & .419 & .121 & 1  \\ \hline
Haberman's-Survival & .05 & .269 & .031 & 5  \\ \hline
Breast-Cancer-Wisconsin & .17 & .46 & .05 & 5 \\ \hline
Statlog(Heart) & .47 & .489 & .0309 & 3  \\ \hline
Seeds & .05 & .189 & .161 & 3  \\ \hline
\end{tabular}%
\end{table}

\begin{table}[h!]
\centering
\caption{Tuned hyperparameters for ChaosFEX$_{RH50L50G}+$kNN.}
\label{table:50L50G:knn }
\centering
\begin{tabular}{llllll}
\hline
Data-set & \begin{tabular}[c]{@{}l@{}}$q$\end{tabular} & \begin{tabular}[c]{@{}l@{}}$b$\end{tabular} & \begin{tabular}[c]{@{}l@{}}$\epsilon$\end{tabular} & $k$ \\ \hline
Iris & .05 & .359 & .221 & 3  \\ \hline
Ionosphere & .099 & .479 & .061 & 3  \\ \hline
Wine & .46 & .469 & .131 & 1  \\ \hline
Bank-Note-Authentication & .09 & .289 & .041 & 1  \\ \hline
Haberman's-Survival & .14 & .489 & .021 & 5 \\ \hline
Breast-Cancer-Wisconsin & .069 & .139 & .041 & 3  \\ \hline
Statlog(Heart) & .18 & .169 & .011 & 3  \\ \hline
Seeds & .05 & .139 & .151 & 3  \\ \hline
\end{tabular}%
\end{table}

\begin{table}[h!]
\centering
\caption{Tuned hyperparameters for ChaosFEX$_{RH75L25G}+$kNN.} 
\label{table:75L25G:knn }
\centering
\begin{tabular}{llllll}
\hline
Data-set & \begin{tabular}[c]{@{}l@{}}$q$\end{tabular} & \begin{tabular}[c]{@{}l@{}}$b$\end{tabular} & \begin{tabular}[c]{@{}l@{}}$\epsilon$\end{tabular} & $k$ \\ \hline
Iris & .15 & .299 & .231 & 5 \\ \hline
Ionosphere & .02 & .219 & .809 & 5  \\ \hline
Wine & .47 & .479 & .131 & 5  \\ \hline
Bank-Note-Authentication & .01 & .259 & .071 & 1  \\ \hline
Haberman's-Survival & .23 & .1 & .011 & 1  \\ \hline
Breast Cancer Wisconsin & .14 & .489 & .021 & 1  \\ \hline
Statlog(Heart) & .13 & .1 & .051 & 5  \\ \hline
Seeds & .05 & .189 & .151 & 3  \\ \hline
\end{tabular}%
\end{table}
\begin{table}[h!]
\centering
\caption{Tuned hyperparameters for ChaosFEX$_{RH25L75G}+$Random Forests.
\label{table:25L75G:Random Forests}}
\centering
\begin{tabular}{lllllll}
\hline
Data-set & \begin{tabular}[c]{@{}l@{}}$q$\end{tabular} & \begin{tabular}[c]{@{}l@{}}$b$\end{tabular} & \begin{tabular}[c]{@{}l@{}}$\epsilon$\end{tabular} & $n\_estimators$ & $max\_depth$  \\ \hline
Iris & .062 & .185 & .298 & 10 & 3 \\ \hline
Ionosphere & .01 & .409 & .051 & 100 & 5 \\ \hline
Wine & .46 & .469 & .141 & 100 & 3  \\ \hline
Bank-Note-Authentication & .36 & .419 & .121 & 10 & 4  \\ \hline
Haberman's-Survival & .05 & .269 & .031 & 10 & 6  \\ \hline
Breast-Cancer-Wisconsin & .17 & .46 & .05 & 10 & 5  \\ \hline
Statlog(Heart) & .47 & .489 & .0309 & 10 & 4  \\ \hline
Seeds & .05 & .189 & .161 & 100 & 4 \\ \hline
\end{tabular}%
\end{table}

\begin{table}[h!]
\centering
\caption{Tuned hyperparameters for ChaosFEX$_{RH50L50G}+$Random Forests. 
\label{table:50L50G:Random Forests}}
\centering
\begin{tabular}{lllllll}
\hline
Data-set & \begin{tabular}[c]{@{}l@{}}$q$\end{tabular} & \begin{tabular}[c]{@{}l@{}}$b$\end{tabular} & \begin{tabular}[c]{@{}l@{}}$\epsilon$\end{tabular} & $n\_estimators$ & $max\_depth$  \\ \hline
Iris & .05 & .359 & .221 & 100 & 3  \\ \hline
Ionosphere & .099 & .479 & .061 & 1000 & 8  \\ \hline
Wine & .46 & .469 & .131 & 1000 & 5  \\ \hline
Bank-Note-Authentication & .09 & .289 & .041 & 100 & 7  \\ \hline
Haberman's-Survival & .14 & .489 & .021 & 1000 & 4 \\ \hline
Breast-Cancer-Wisconsin & .069 & .139 & .041 & 10 & 8  \\ \hline
Statlog(Heart) & .18 & .169 & .011 & 100 & 5  \\ \hline
Seeds & .05 & .139 & .151 & 100 & 4  \\ \hline
\end{tabular}%
\end{table}
\begin{table}[h!]
\centering
\caption{Tuned hyperparameters for ChaosFEX$_{RH75L25G}+$Random Forests.
\label{table:75L25G:Random Forests}}
\centering
\begin{tabular}{lllllll}
\hline
Data-set & \begin{tabular}[c]{@{}l@{}}$q$\end{tabular} & \begin{tabular}[c]{@{}l@{}}$b$\end{tabular} & \begin{tabular}[c]{@{}l@{}}$\epsilon$\end{tabular} & $n\_estimators$ & $max\_depth$ \\ \hline
Iris & .15 & .299 & .231 & 10 & 2  \\ \hline
Ionosphere & .02 & .219 & .809 & 10 & 5 \\ \hline
Wine & .47 & .479 & .131 & 100 & 6  \\ \hline
Bank-Note-Authentication & .01 & .259 & .071 & 100 & 7  \\ \hline
Haberman's-Survival & .23 & .1 & .011 & 10 & 5  \\ \hline
Breast-Cancer-Wisconsin & .14 & .489 & .021 & 10 & 10  \\ \hline
Statlog(Heart) & .13 & .1 & .051 & 10 & 2  \\ \hline
Seeds & .05 & .189 & .151 & 100 & 4  \\ \hline
\end{tabular}%
\end{table}

\begin{table}[h!]
\centering
\caption{Tuned hyperparameters for  ChaosFEX$_{RH25L75G}$ for {\it FSDD}.}
\begin{tabular}{|l|l|}
\hline
\textbf{Hyper-parameter} & \textbf{Tuned Value} \\ \hline
$q$                       & .086               \\ \hline
$b$                       & .303                \\ \hline
$\epsilon$              & .055                \\ \hline
\end{tabular}
\label{table:FSSD 25L75G Hyperparameters}
\end{table}
\begin{table}[h!]
\centering
\caption{Tuned hyperparameters for  ChaosFEX$_{RH50L50G}$ for {\it FSDD}.}
\begin{tabular}{|l|l|}
\hline
\textbf{Hyper-parameter} & \textbf{Tuned Value} \\ \hline
$q$                       & .106               \\ \hline
$b$                       & .032                \\ \hline
$\epsilon$              & .104                \\ \hline
\end{tabular}

\label{table:FSSD 50L50GHyperparameters}
\end{table}

\begin{table}[h!]
\centering
\caption{Tuned hyperparameters for  ChaosFEX$_{RH75L25G}$ for {\it FSDD}.}
\begin{tabular}{|l|l|}
\hline
\textbf{Hyper-parameter} & \textbf{Tuned Value} \\ \hline
$q$                       & .40               \\ \hline
$b$                       & .20                \\ \hline
$\epsilon$              & .15                \\ \hline
\end{tabular}
\label{table:FSSD 75L25G Hyperparameters}
\end{table}

\begin{table}[!h]
\centering
\caption{Tuned hyperparameters for various classifiers for {\it FSDD}.}
\label{table:hyperparameters_FSDD_OtherClassifiers}
\begin{tabular}{llll}
\hline
\textbf{Classifiers} & ChaosFEX$_{RH25L75G}$ & ChaosFEX$_{RH50L50G}$ & ChaosFEX$_{RH75L25G}$ \\ \hline
AdaBoost & $n\_estimators = 50$ & $n\_estimators =1$ & $n\_estimators =10$ \\ \hline
\multirow{3}{*}{Decision Trees} & $min\_samples\_leaf =1$ & $min\_samples\_leaf =1$ & $min\_samples\_leaf =4$ \\ \cline{2-4} 
 & $max\_depth=2$ & $max\_depth=2$ & $max\_depth=7$ \\ \cline{2-4} 
 & $ccp\_alpha=0.0074$ & $ccp\_alpha=0.0$ & $ccp\_alpha=0.00723$ \\ \hline
k-NN & $k=3$ & $k=5$ & $k=1$ \\ \hline
\multirow{2}{*}{Random Forests} & $n\_estimators = 2$ & $n\_estimators =2$ & $n\_estimators =2$ \\ \cline{2-4} 
 & $max\_depth=1000$ & $max\_depth=1000$ & $max\_depth=10$ \\ \hline
\end{tabular}%
\end{table}
\begin{table}[h!]
\caption{Parameters tuned for various $ChaosFEX_{RH25L75G}$ structures considered for analysis with debris-urban dataset}
\label{table:25L75Gparameters debrisurban}
\centering
\begin{tabular}{lll}
\hline
Algorithm & Hyper Parameters \\ \hline
$ChaosFEX_{RH25L75G}$ & \begin{tabular}[c]{@{}l@{}}$q$=.3649 \\ $b$=.430\\ $\epsilon$=.259\end{tabular} \\ \hline
$ChaosFEX_{RH25L75G}$+SVM & \begin{tabular}[c]{@{}l@{}}$q$=.3649 \\ $b$=.430\\ $\epsilon$=.259\end{tabular} \\ \hline
$ChaosFEX_{RH25L75G}$+k-NN & \begin{tabular}[c]{@{}l@{}}$q$=.3649 \\ $b$=.430\\ $\epsilon$=.259\\ k=3\end{tabular}  \\ \hline
$ChaosFEX_{RH25L75G}$+AdaBoost & \begin{tabular}[c]{@{}l@{}}$q$=.3649 \\ $b$=.430\\ $\epsilon$=.259\\ $n\_estimator$=1\end{tabular}  \\ \hline
$ChaosFEX_{RH25L75G}$+Decision Tree & \begin{tabular}[c]{@{}l@{}}$q$=.3649 \\ $b$=.430\\ $\epsilon$=.259\\ $min\_samples\_leaf$ = 1\\ $random\_state$ = 42\\ $max\_depth$ = 6\\ $ccp\_alpha$ = 0\end{tabular}  \\ \hline
$ChaosFEX_{RH25L75G}$+GNB & \begin{tabular}[c]{@{}l@{}}$q$=.3649 \\ $b$=.430\\ $\epsilon$=.259\end{tabular}  \\ \hline
$ChaosFEX_{RH25L75G}$+RF & \begin{tabular}[c]{@{}l@{}}$q$=.3649 \\ $b$=.430\\ $\epsilon$=.259\\ $n\_estimators = 1000$, \\ $max\_depth $= 8\end{tabular}  \\ \hline
\end{tabular}%
\end{table}

\begin{table}[h!]
\caption{Parameters tuned for various $ChaosFEX_{RH50L50G}$ structures considered for analysis with debris-urban dataset}
\label{table50L50Gparametersdebrisurban}
\centering
\begin{tabular}{ll}
\hline
Algorithm & Hyper Parameters \\ \hline
$ChaosFEX_{RHNL50L50G}$ & \begin{tabular}[c]{@{}l@{}}$q$ = .121\\ $b$ = .0041\\ $\epsilon$ = .015\end{tabular} \\ \hline
$ChaosFEX_{RHNL50L50G}$+SVM & \begin{tabular}[c]{@{}l@{}}$q$ = .121\\ $b$ = .0041\\ $\epsilon$ = .015\end{tabular} \\ \hline
$ChaosFEX_{RHNL50L50G}$+k-NN & \begin{tabular}[c]{@{}l@{}}$q$ = .121\\ $b$ = .0041\\ $\epsilon$ = .015\\ $k=1$\end{tabular} \\ \hline
$ChaosFEX_{RHNL50L50G}$+Decision Tree & \begin{tabular}[c]{@{}l@{}}$q$ = .121\\ $b$ = .0041\\ $\epsilon$ = .015\\ $min\_samples\_leaf$ = 1\\ $max\_depth$= 4\\ $ccp\_alpha$ = 0\end{tabular} \\ \hline
$ChaosFEX_{RHNL50L50G}$+GNB & \begin{tabular}[c]{@{}l@{}}$q$ = .121\\ $b$ = .0041\\ $\epsilon$ = .015\end{tabular} \\ \hline
$ChaosFEX_{RHNL50L50G}$+AdaBoost & \begin{tabular}[c]{@{}l@{}}$q$ = .121\\ $b$ = .0041\\ $\epsilon$ = .015\\ $n\_estimators$=3\end{tabular} \\ \hline
$ChaosFEX_{RHNL50L50G}$+Random Forest & \begin{tabular}[c]{@{}l@{}}$q$ = .121\\ $b$ = .0041\\ $\epsilon$ = .015\\ $n\_estimators$ = 1000\\ $max\_depth$ = 5\end{tabular} \\ \hline
\end{tabular}%
\end{table}

\begin{table}[h!]
\caption{Parameters tuned for various $ChaosFEX_{RH75L25G}$ structures considered for analysis with debris-urban dataset}
\label{table75L25Gparametersdebrisurban}
\centering
\begin{tabular}{ll}
\hline
Algorithm & Hyper Parameters \\ \hline
$ChaosFEX_{RHNL75L25G}$ & \begin{tabular}[c]{@{}l@{}}$q$= .491\\ $b$= .010\\ $\epsilon$= .0856\end{tabular} \\ \hline
$ChaosFEX_{RHNL75L25G}$+SVM & \begin{tabular}[c]{@{}l@{}}$q$= .491\\ $b$= .010\\ $\epsilon$= .0856\end{tabular} \\ \hline
$ChaosFEX_{RHNL75L25G}$+k-NN & \begin{tabular}[c]{@{}l@{}}$q$= .491\\ $b$= .010\\ $\epsilon$= .0856\\ $k=5$\end{tabular} \\ \hline
$ChaosFEX_{RHNL75L25G}$+Decision Tree & \begin{tabular}[c]{@{}l@{}}$q$= .491\\ $b$= .010\\ $\epsilon$= .0856\\ $random\_state$=42\\ $min\_samples\_leaf$ = 1\\ $max\_depth$= 4\\ $ccp\_alpha$ = 0\end{tabular} \\ \hline
$ChaosFEX_{RHNL75L25G}$+GNB & \begin{tabular}[c]{@{}l@{}}$q$= .491\\ $b$= .010\\ $\epsilon$= .0856\end{tabular} \\ \hline
$ChaosFEX_{RHNL75L25G}$+AdaBoost & \begin{tabular}[c]{@{}l@{}}$q$= .491\\ $b$= .010\\ $\epsilon$= .0856\\ $n\_estimators$=100\end{tabular} \\ \hline
$ChaosFEX_{RHNL75L25G}$+Random Forest & \begin{tabular}[c]{@{}l@{}}$q$= .491\\ $b$= .010\\ $\epsilon$= .0856\\ $n\_estimators$ = 100\\ $max\_depth$ = 4\end{tabular} \\ \hline
\end{tabular}%
\end{table}
\begin{table}[]
\caption{Parameters tuned for various $ChaosFEX_{RH25L50G}$ structures considered
for analysis with Brain Tumor dataset}
\label{table:parameter25L75GBrainTumor}
\centering
\begin{tabular}{ll}
\hline
Algorithm & Parameters \\ \hline
$ChaosFEX_{RH25L75G}$ & \begin{tabular}[c]{@{}l@{}}$q$=.01\\ $b$=.36\\ $\epsilon$=.090\end{tabular} \\ \hline
$ChaosFEX_{RH25L75G}$+SVM & \begin{tabular}[c]{@{}l@{}}$q$=.01 \\ $b$=.36\\ $\epsilon$=.090\end{tabular} \\ \hline
$ChaosFEX_{RH25L75G}$+k-NN & \begin{tabular}[c]{@{}l@{}}$q$=.01\\ $b$=.430\\ $\epsilon$=.230\\ $k$=3\end{tabular} \\ \hline
$ChaosFEX_{RH25L75G}$+Decision Tree & \begin{tabular}[c]{@{}l@{}}$q$=.01\\ $b$=.430\\ $\epsilon$=.230\\ $min\_samples\_leaf$ = 1\\ $max\_depth$ = 6\\ $ccp\_alpha$ = 0\end{tabular} \\ \hline
$ChaosFEX_{RH25L75G}$+GNB & \begin{tabular}[c]{@{}l@{}}$q$=.01\\ $b$=.430\\ $\epsilon$=.230\end{tabular} \\ \hline
$ChaosFEX_{RH25L75G}$+AdaBoost & \begin{tabular}[c]{@{}l@{}}$q$=.01\\ $b$=.430\\ $\epsilon$=.230\\ $n\_estimator$= 1000\end{tabular} \\ \hline
$ChaosFEX_{RH25L75G}$+Random Forest & \begin{tabular}[c]{@{}l@{}}$q$=.01\\ $b$=.430\\ $\epsilon$=0.230\\ $n\_estimator$= 1000\\ $max\_depth$ = 8\end{tabular} \\ \hline
\end{tabular}%
\end{table}

\begin{table}[!h]
\caption{Parameters tuned for various $ChaosFEX_{RH50L50G}$ structures considered
for analysis with Brain Tumor dataset}
\label{table:hyperparameters50L50Gbraintumor}
\centering
\begin{tabular}{ll}
\hline
Algorithm & Hyper Parameters \\ \hline
$ChaosFEX_{RH50L50G}$ & \begin{tabular}[c]{@{}l@{}}$q$ = .094\\ $b$ = .0065\\ $\epsilon$ = .0092\end{tabular} \\ \hline
$ChaosFEX_{RH50L50G}$+SVM & \begin{tabular}[c]{@{}l@{}}$q$ = .094\\ $b$ = .0065\\ $\epsilon$ = .0092\end{tabular} \\ \hline
$ChaosFEX_{RH50L50G}$+k-NN & \begin{tabular}[c]{@{}l@{}}$q$ = .094\\ $b$ = .0065\\ $\epsilon$ = .0092\\ $k$=1\end{tabular} \\ \hline
$ChaosFEX_{RH50L50G}$+Decision Tree & \begin{tabular}[c]{@{}l@{}}$q$ = .094\\ $b$ = .0065\\ $\epsilon$ = .0092\\ $min\_samples\_leaf$ = 1\\ $max\_depth$ = 4\\ $ccp\_alpha$ = 0\end{tabular} \\ \hline
$ChaosFEX_{RH50L50G}$+GNB & \begin{tabular}[c]{@{}l@{}}$q$ = .094\\ $b$ = .0065\\ $\epsilon$ = .0092\end{tabular} \\ \hline
$ChaosFEX_{RH50L50G}$+AdaBoost & \begin{tabular}[c]{@{}l@{}}$q$ = .094\\ $b$ = .0065\\ $\epsilon$ = .0092\\ $n\_estimator$= 3\end{tabular} \\ \hline
$ChaosFEX_{RH50L50G}$+Random Forest & \begin{tabular}[c]{@{}l@{}}$q$ = .094\\ $b$ = .0065\\ $\epsilon$ = .0092\\ $n\_estimator$= 1000\\ $max\_depth$ = 5\end{tabular} \\ \hline
\end{tabular}%
\end{table}
\begin{table}[]
\caption{Parameters tuned for various $ChaosFEX_{RH75L25G}$ structures considered
for analysis with Brain Tumor dataset}
\label{table:parameters75L25GBT}
\centering
\begin{tabular}{ll}
\hline
Algorithm & Parameters \\ \hline
$ChaosFEX_{RH75L25G}$ & \begin{tabular}[c]{@{}l@{}}$q$ = .156\\ $b$ = .107\\ $\epsilon$= .059\end{tabular} \\ \hline
$ChaosFEX_{RH75L25G}$+SVM & \begin{tabular}[c]{@{}l@{}}$q$ = .156\\ $b$ = .107\\ $\epsilon$= .059\end{tabular} \\ \hline
$ChaosFEX_{RH75L25G}$+k-NN & \begin{tabular}[c]{@{}l@{}}$q$ = .156\\ $b$ = .107\\ $\epsilon$= .059\\ $k$=5\end{tabular} \\ \hline
$ChaosFEX_{RH75L25G}$+Decision Tree & \begin{tabular}[c]{@{}l@{}}$q$ = .156\\ $b$ = .107\\ $\epsilon$= .059\\ $min\_samples\_leaf$ = 10\\ $max\_depth$ = 4\\ $ccp\_alpha$ = 0.0\end{tabular} \\ \hline
$ChaosFEX_{RH75L25G}$+GNB & \begin{tabular}[c]{@{}l@{}}$q$ = .156\\ $b$ = .107\\ $\epsilon$= .059\end{tabular} \\ \hline
$ChaosFEX_{RH75L25G}$+AdaBoost & \begin{tabular}[c]{@{}l@{}}$q$ = .156\\ $b$ = .107\\ $\epsilon$= .059\\ $n\_estimator$= 10\end{tabular} \\ \hline
$ChaosFEX_{RH75L25G}$+Random Forest & \begin{tabular}[c]{@{}l@{}}$q$ = .156\\ $b$ = .107\\ $\epsilon$= .059\\ $n\_estimator$= 100\\ $max\_depth$ = 4\end{tabular} \\ \hline
\end{tabular}%
\end{table}




\end{document}